\PassOptionsToPackage{full}{textcomp}
\documentclass[conference]{IEEEtran}
\IEEEoverridecommandlockouts

\usepackage[a4paper, margin=2cm]{geometry}
\usepackage{amsmath,mathtools}
\usepackage[ruled,vlined]{algorithm2e}
\usepackage{hyperref}
\usepackage{graphicx}
\usepackage[caption=false]{subfig}
\usepackage{tabularray}
\usepackage{booktabs}
\usepackage{framed}
\usepackage{cleveref}
\usepackage[dvipsnames]{xcolor}
\usepackage{enumitem}
\usepackage[noadjust]{cite}

\AtBeginEnvironment{appendices}{\crefalias{section}{appendix}}

\UseTblrLibrary{booktabs}
\UseTblrLibrary{diagbox}

\newcommand{\etal}{\emph{et al.}\xspace}
\newcommand{\eg}{\emph{e.g.,}\xspace}
\newcommand{\ie}{\emph{i.e.,}\xspace}

\newcommand{\remove}[1]{}
\newcommand{\reduce}[1]{}

\begin{document}

\title{
Balancing Privacy--Quality--Efficiency in Federated Learning through Round-Based Interleaving of Protection Techniques
}

\author{\IEEEauthorblockN{Yenan Wang}
\IEEEauthorblockA{Computer Science and Engineering \\
Chalmers University of Technology\\
Gothenburg, Sweden 
}
\and
\IEEEauthorblockN{Carla Fabiana Chiasserini}
\IEEEauthorblockA{Computer Science and Engineering \\
Chalmers University of Technology\\
Gothenburg, Sweden
}
\and
\IEEEauthorblockN{Elad Michael Schiller}
\IEEEauthorblockA{Computer Science and Engineering \\
Chalmers University of Technology \\
Gothenburg, Sweden} 
}

\maketitle
\begin{abstract}
In federated learning (FL), balancing privacy protection, learning quality, and efficiency remains a challenge.
Privacy protection mechanisms, such as Differential Privacy (DP), degrade learning quality, or, as in the case of Homomorphic Encryption (HE), incur substantial system overhead.
To address this, we propose Alt-FL, a privacy-preserving FL framework that combines DP, HE, and synthetic data via a novel round-based interleaving strategy.
Alt-FL introduces three new methods, Privacy Interleaving (PI), Synthetic Interleaving with DP (SI/DP), and Synthetic Interleaving with HE (SI/HE), that enable flexible quality--efficiency trade-offs while providing  privacy~protection.

We systematically evaluate Alt-FL against representative reconstruction attacks, including Deep Leakage from Gradients, Inverting Gradients, When the Curious Abandon Honesty, and Robbing the Fed, using a LeNet-5 model on CIFAR-10 and Fashion-MNIST.
To enable fair comparison between DP- and HE-based defenses, we introduce a new attacker-centric framework that compares empirical attack success rates across the three proposed interleaving methods.
Our results show that, for the studied attacker  model and dataset, PI achieves the most balanced trade-offs at high privacy protection levels, while DP-based methods are preferable at intermediate privacy requirements.
We also discuss how such results can be the basis for selecting privacy-preserving FL methods under varying privacy and resource constraints.
\end{abstract}

\section{Introduction}

Federated Learning (FL)~\cite{DBLP:conf/aistats/McMahanMRHA17,DBLP:journals/tist/YangLCT19} enables collaborative model training by sharing only model updates with a central server.
FL has thus been widely adopted in privacy-critical domains  such as healthcare, banking, and smart cities.
However, these applications must also operate under strict communication and computational constraints.

Although raw data are not transmitted, repeated update transmissions can still introduce overhead and security vulnerabilities~\cite{DBLP:series/lncs/Zhu020,DBLP:conf/nips/GeipingBD020,DBLP:conf/eurosp/BoenischDSSSP23,DBLP:conf/iclr/FowlGCGG22}.
Zhu \etal~\cite{DBLP:series/lncs/Zhu020} showed that shared gradients enable data reconstruction via the \emph{Deep Leakage from Gradients} (DLG) attack, compromising client-level privacy.
Geiping \etal~\cite{DBLP:conf/nips/GeipingBD020} improved this attack with \emph{Inverting Gradients} (Inverting), achieving higher-fidelity reconstructions.
Under stronger attacker models, \emph{When the Curious Abandon Honesty} (CAH)~\cite{DBLP:conf/eurosp/BoenischDSSSP23} and \emph{Robbing the Fed} (RTF)~\cite{DBLP:conf/iclr/FowlGCGG22} can even achieve near-perfect reconstruction.

\subsection{Open Issues and Proposed Solution}

\begin{figure*}[htbp]
    \centering
    \includegraphics[width=\linewidth]{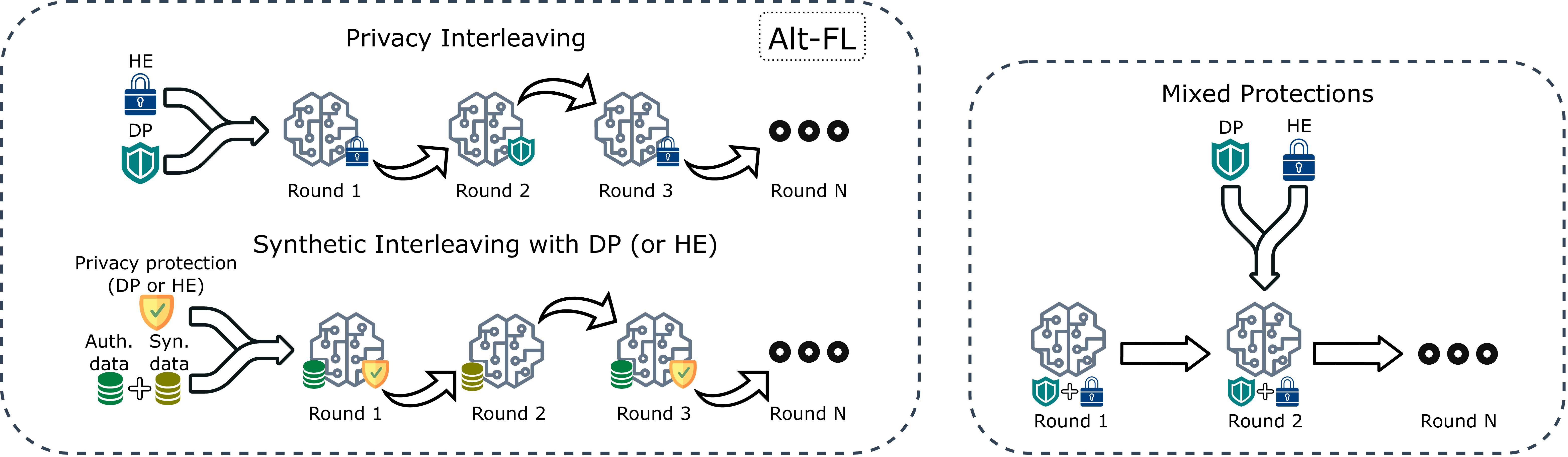}
    \caption{Overview of the proposed Alt-FL schemes (left) vs. the baseline Mixed Protections (right).}
    \label{fig:conceptFigure}
\end{figure*}

Privacy protection mechanisms involve inherent trade-offs between privacy, learning quality, and efficiency.
Namely, Differential Privacy (DP)~\cite{DBLP:conf/ccs/AbadiCGMMT016} degrades accuracy through noise injection, while Homomorphic Encryption (HE)~\cite{DBLP:conf/asiacrypt/CheonKKS17} incurs high communication and computational overhead.
Despite extensive research in FL systems, efficiently balancing \emph{all} these trade-offs remains an open challenge.

To tackle this challenge, we propose the Alternating FL (Alt-FL) framework, which interleaves DP, HE, and synthetic data to balance the privacy--quality--efficiency trade-off (see \Cref{fig:conceptFigure}). 
Specifically, we develop and use:

\noindent $\bullet$ \emph{Combined privacy protection mechanism.} 
We leverage the combination of DP and Selective Homomorphic Encryption (S-HE)~\cite{DBLP:journals/corr/abs-2303-10837}, with the latter applying HE only to the most sensitive information to reduce overheads.

\noindent $\bullet$ 
\emph{Interleaved privacy protections.} 
We propose a novel interleaving method that alternates between \emph{DP rounds} and \emph{HE rounds}, referred to as \emph{Privacy Interleaving} (PI), as illustrated in \Cref{fig:conceptFigure}.
PI reduces DP-induced quality loss and S-HE overhead while maintaining privacy protection in each round.
We control the proportion of DP and HE rounds using a tunable interleaving ratio $\rho$.

\noindent $\bullet$ 
\emph{Interleaved authentic and synthetic data.}  
We also propose new interleaving methods that alternate between \emph{authentic} and \emph{synthetic} rounds.
These methods, denoted as \emph{Synthetic Interleaving with DP} (SI/DP) and \emph{Synthetic Interleaving with HE} (SI/HE), apply DP or S-HE protection to protect authentic data during authentic rounds, while training on synthetic data without protection during synthetic rounds (see \Cref{fig:conceptFigure}).
As in PI, the interleaving ratio $\rho$ controls the proportion of synthetic to authentic~rounds.

\noindent $\bullet$ \emph{Unified empirical analysis of privacy--quality--efficiency trade-offs.}
We empirically evaluate privacy, learning quality, and efficiency under the proposed methods (PI, SI/DP, and SI/HE).
For privacy, we focus on protecting client-level data by measuring empirical attack success rates under state-of-the-art reconstruction attacks.
We then compare the ability of the  proposed methods to defend against privacy attacks with a baseline, named \emph{Mixed Protections} (MP), which applies DP and HE in every training round.

\subsection{Our Contribution}

\noindent $\bullet$
We present a novel framework to empirically quantify \emph{privacy protection levels} using attacker-centric metrics.
By jointly comparing DP and S-HE using combinations of noise multipliers $\sigma$ (used in DP) and encryption ratios $\eta$ (used in S-HE),  we define these privacy protection levels based on empirical attack success rates (\Cref{sec:privacy_levels}).

\noindent $\bullet$
We propose three tunable methods (PI, SI/DP, and SI/HE) based on a novel interleaving approach for combining DP, HE, and synthetic data to balance learning quality and efficiency while defending against the studied attacks (\Cref{sec:rq1_integration}).

\noindent $\bullet$
We systematically evaluate these methods across varying privacy protection levels and data distribution settings (\Crefrange{sec:rq1_integration}{sec:privacy_vs_system_costs}).
We show how to identify methods with favorable privacy--quality--efficiency trade-offs for the studied dataset and model by addressing the following research questions. 

\indent  \emph{RQ1 Privacy Preservation:} To what extent do the combined techniques preserve privacy against data reconstruction attacks?
For a given privacy protection level and data distributions, which techniques provide suitable defenses?

\indent  \emph{RQ2 System costs vs. Learning Quality:} What are the performance and system costs (e.g., convergence time, communication cost, and computational cost) associated with privacy preservation?
For a given privacy defense level and data distribution, which techniques offer the best trade-offs between system costs and learning quality?

\indent  \emph{RQ3 Privacy Preservation vs. System Costs:} Under a required accuracy level, how do combinations of DP, S-HE, and synthetic data perform in terms of attack success rates and system costs?
For a given accuracy constraint, privacy protection level, and data distribution, which techniques offer the best trade-offs between privacy preservation and system costs?

\noindent $\bullet$ Our empirical analysis provides guidance for selecting cost-effective privacy-preserving methods evaluated on the LeNet-5 model~\cite{DBLP:journals/pieee/LeCunBBH98} with the CIFAR-10~\cite{krizhevsky2009learning} and  Fashion-MNIST dataset~\cite{DBLP:journals/corr/abs-1708-07747}.
Our results in \Cref{fig:methods_high_level_objective} answer RQ1--RQ3 and show that the optimal method depends on required privacy and cost constraints, 
specifically: 

\indent -- 
At the highest privacy protection levels, PI provides the most balanced privacy--quality-efficiency trade-off;

\indent --
At intermediate privacy protection levels, DP-based approaches (SI/DP and DP-only) achieve more favorable trade-offs;

\indent --
At the weakest privacy protection levels, HE-based approaches (MP, SI/HE, and HE-only) become necessary to ensure protection;

\smallskip
To allow for the reproducibility of our results and foster further development, we pledge to release the implementation of our solution upon acceptance of the paper.

\section{Related Work and Motivation}

The challenges of balancing privacy, learning quality, and efficiency have been formalized by Zhang \etal~\cite{DBLP:journals/tist/ZhangKCFY23}, who proposed a theoretical framework that quantifies these trade-offs in FL.
Under this framework, they proved that no mechanism can simultaneously optimize all three of these objectives.
Hence, our study focuses on identifying methods achieving the most favorable trade-offs.

Prior work studies how data quality affects convergence~\cite{DBLP:conf/icml/KarimireddyKMRS20,DBLP:conf/iclr/Huang0CS24}.
Other studies focus on synthetic data generation~\cite{DBLP:journals/corr/abs-2302-04062}, which can improve data diversity but incur additional computational cost.
Synthetic data are also used for data augmentation~\cite{MIR-2022-08-256}, though training exclusively on synthetic data degrades model quality.
Moreover, na\"ively generated synthetic data may leak sensitive information~\cite{DBLP:conf/aistats/BreugelSQS23,DBLP:journals/jbi/ZhangYM22}.
In our work, we study how privacy-preserving synthetic data can be combined with existing privacy protection mechanisms.

The FL landscape is threatened by various privacy attacks, including membership~\cite{DBLP:journals/csur/BaiHYLWX25}, property~\cite{DBLP:journals/tdsc/WangHSWXR23}, and source inference attacks~\cite{DBLP:journals/tdsc/HuZSSCD24} that can infer training participation, properties, and data sources, respectively.
Recent work also shows that private attributes can be inferred from protected data via black-box probing attacks~\cite{DBLP:conf/icml/Chen0HHA25}.
Beyond inference-based attacks, data reconstruction attacks~\cite{DBLP:series/lncs/Zhu020,DBLP:conf/nips/GeipingBD020,DBLP:conf/eurosp/BoenischDSSSP23,DBLP:conf/iclr/FowlGCGG22} can directly recover raw training data from model updates.
In this work, we focus on data reconstruction attacks.

\noindent \textbf{Data Reconstruction Attacks.}
We consider four representative data reconstruction attacks.
The {\em DLG attack} by Zhu \etal~\cite{DBLP:series/lncs/Zhu020} formulates reconstruction as an optimization problem, iteratively matching the gradients computed on dummy data to the observed gradients.
Geiping \etal~\cite{DBLP:conf/nips/GeipingBD020} extend this idea with the {\em Inverting attack}, improving stability and reconstruction fidelity via a modified loss function and regularization.
Geiping \etal~\cite{DBLP:conf/nips/GeipingBD020} further show that inputs to fully connected layers can be analytically recovered from gradients.
The {\em CAH attack}~\cite{DBLP:conf/eurosp/BoenischDSSSP23} exploits this via trap weights to amplify gradient leakage, while the {\em RTF attack}~\cite{DBLP:conf/iclr/FowlGCGG22} inserts an imprint module to encode client data, enabling near-exact reconstructions.
Notably, although \emph{See Through Gradients} by Yin \etal~\cite{DBLP:conf/cvpr/YinMVAKM21} reports state-of-the-art performance among optimization-based attacks, we exclude it from our evaluation as there is no publicly available implementation that achieves their reported performance.

\noindent {\bf Privacy Preservation in FL.} Since the introduction of FedAvg~\cite{DBLP:conf/aistats/McMahanMRHA17}, several solutions have been proposed to prevent privacy leakage from model updates.
For instance, Bonawitz \etal~\cite{DBLP:journals/corr/BonawitzIKMMPRS16} developed a secure aggregation protocol using double-masking to protect against honest-but-curious servers.
However, secure aggregation requires complex client-to-client coordination.
We therefore omit it to focus on methods compatible with standard centralized FL, specifically DP and HE.

DP in FL is commonly implemented via DP-SGD~\cite{DBLP:conf/ccs/AbadiCGMMT016}, which adds calibrated noise to model gradients.
Although it provides formal guarantees, empirical studies show a trade-off between privacy and learning quality~\cite{DBLP:conf/ccs/AbadiCGMMT016,DBLP:conf/uss/YangHYGC23,DBLP:journals/corr/abs-2510-19934}.
PrivateFL~\cite{DBLP:conf/uss/YangHYGC23} mitigates this via personalized data transformations to counteract the heterogeneity introduced by DP, and
Li \etal~\cite{DBLP:journals/corr/abs-2510-19934} utilize the $f$-differential privacy framework for a more refined accounting to improve model accuracy in decentralized settings.
We adopt DP-SGD as the de facto DP standard in our implementation.

HE enables computation on encrypted data without DP-induced quality loss but incurs high computational and communication overhead~\cite{DBLP:journals/fi/FangQ21,DBLP:journals/corr/abs-2303-10837}.
To mitigate this, Cheon \etal~\cite{DBLP:conf/asiacrypt/CheonKKS17} introduced a scheme with a reduced ciphertext size.
Jin \etal~\cite{DBLP:journals/corr/abs-2303-10837} leverage this in FedML-HE through Selective Homomorphic Encryption (S-HE) with a tunable encryption ratio.
Fang \etal~\cite{DBLP:journals/fi/FangQ21} adopt  an improved Paillier-based HE to reduce training time but do not account for the communication cost.
We therefore build on S-HE to further reduce communication overhead.

Several hybrid approaches have explored the combination of DP and HE.
S\'{e}bert \etal~\cite{DBLP:journals/corr/abs-2205-04330} propose a method that applies DP and HE simultaneously to ensure robust privacy.
We leverage this concept as our baseline MP and compare it to our novel interleaving-based methods.

Recently, Korkmaz and Rao~\cite{DBLP:journals/corr/abs-2501-12911} propose a variation of MP that incorporates cryptographic bitwise scrambling to enhance privacy protection.
In the discussion of their scrambling technique, the authors refer readers to reference [24] in \cite{DBLP:journals/corr/abs-2501-12911}.
However, no verifiable record of this reference could be located in standard bibliographic~sources.

Liu and Gupta~\cite{DBLP:conf/icissp/LiuG24} introduce Aero, a variation on the MP method we consider, which demonstrates efficient FL at large-scale, with up to $10^9$ clients.
Aero demonstrates that MP-style protection can be extended to much larger-scale deployments using per-round random committee election, collaborative DP noise generation, central DP, and additive HE.
We implement MP in a conventional client--server FL setting using local DP and a more general HE scheme, specifically an efficient variant of full HE~\cite{DBLP:journals/corr/abs-2303-10837}.

Other hybrid approaches that have explored the combination of  DP and HE include Tang \etal~\cite{DBLP:journals/corr/abs-1812-02292} and Negoya \etal~\cite{negoya2025}.
Tang \etal~\cite{DBLP:journals/corr/abs-1812-02292} combine DP and HE but rely on interpretable tabular features for encryption or perturbation with noise, limiting applicability to high-dimensional data.
Negoya \etal~\cite{negoya2025} propose a flexible FL framework where clients choose adaptively between DP or HE based on their individual resource constraints.
In contrast, our approach systematically balances accuracy and system costs independent of client-specific resource conditions.

Integrating DP with other advanced cryptographic schemes has been explored.
HybridAlpha~\cite{DBLP:conf/ccs/XuBZAL19} combines DP with Multi-Input Functional Encryption (MIFE).
MIFE enables the server to compute aggregated model results without accessing individual updates.
However, the aggregated model is in plaintext and remains susceptible to stronger reconstruction attacks~\cite{DBLP:conf/iclr/FowlGCGG22,DBLP:conf/eurosp/BoenischDSSSP23}.
To mitigate this, HybridAlpha combines DP with MIFE, which may impact learning quality.
In contrast, our work studies new interleaving approaches, which balance trade-offs among privacy, learning quality, and system cost, rather than focusing primarily on reducing system costs as in HybridAlpha.
Furthermore, HybridAlpha relies on a Trusted Third Party (TTP), which limits decentralization~\cite{DBLP:journals/percom/TsouvalasMBOFM25}.

To overcome the TTP limitation, Tsouvalas \etal~\cite{DBLP:journals/percom/TsouvalasMBOFM25} proposed EncCluster, which employs Decentralized Multi-Client Functional Encryption (DMCFE) combined with weight clustering.
While this reduces trust assumptions, the aggregated model remains exposed to the server, making it vulnerable under stronger adversarial models, which we consider (see \Cref{sec:methods})~\cite{DBLP:conf/iclr/FowlGCGG22,DBLP:conf/eurosp/BoenischDSSSP23}.
Moreover, although the authors intended to release the code, it is not publicly available at the time of writing.
We therefore focus our comparison on reproducible frameworks.

\begin{figure}[tbp]
    \centering
    \subfloat[]{
    \centering
    \includegraphics[width=0.97\linewidth]{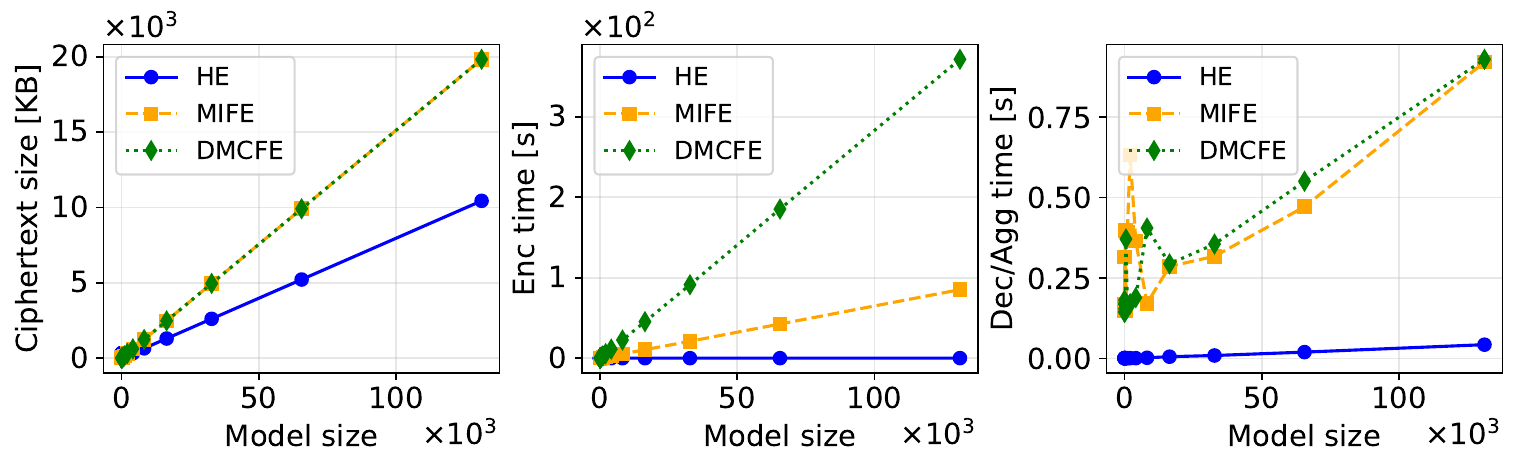}
    }\\[0.1cm]
   \subfloat[]{
    \centering
    \includegraphics[width=0.97\linewidth]{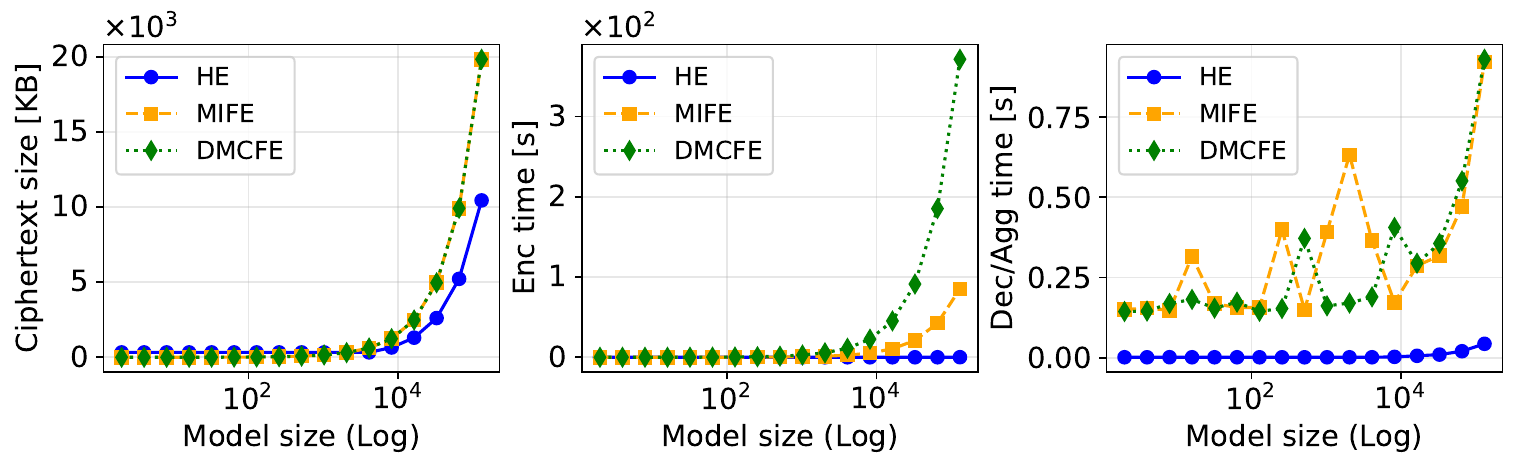}
    } 

    \caption{Comparison of ciphertext size (left), encryption time (middle), and combined decryption and aggregation time (right) across HE, MIFE, and DMCFE schemes. (a) linear-scale x-axis. (b) logarithmic-scale x-axis.}
    \label{fig:fe_vs_he_performance}
\end{figure}

\noindent \textbf{Comparison of FE and HE.} 
Our related conference paper~\cite{DBLP:conf/lanman/WangCS25}, proposed and studied the SI/HE method.
Our work greatly extends  \cite{DBLP:conf/lanman/WangCS25}, as we now propose SI/DP and PI and compare them to the baseline method MP. 
We also focus on DP and S-HE (rather than MIFE or DMCFE) due to the following  observation based on the empirical analysis we performed using the open-source TenSEAL~\cite{tenseal2021} and PyMIFE~\cite{PyMIFE_2024} libraries. 
For LeNet-5~\cite{DBLP:journals/pieee/LeCunBBH98}, the results in \Cref{fig:fe_vs_he_performance} show that HE achieves better performance compared to MIFE and DMCFE.
We therefore adopt HE due to its superior efficiency in our FL setting.

Specifically, we compare the performance of functional encryption (FE) and homomorphic encryption (HE) in a FL setting.
For HE, we use the TenSEAL Python library~\cite{tenseal2021} and evaluate the CKKS encryption scheme, which is the scheme used throughout this work.
For FE, we consider two classes of schemes that have been studied in FL contexts~\cite{DBLP:conf/ccs/XuBZAL19,DBLP:journals/percom/TsouvalasMBOFM25}: MIFE and DMCFE.
We use the FE implementations provided in~\cite{PyMIFE_2024}.

To simplify the FL setting, we assume that the model parameters are already vectorized and represented as a single vector of numerical values.
The setup involves three clients, each encrypting an identical model parameter vector.
We vary the vector length over the sequence $(2, 2^2, 2^3, \dots, 2^{17})$, with entries sampled at random.
After encryption, the encrypted models are aggregated, and the final result is decrypted.
Although HE-based aggregation does not require decryption to obtain the aggregated encrypted model, we include a decryption step to enable a consistent comparison across schemes.

All experiments are executed on a single machine within a single script, simulating the FL workflow.
For HE, aggregation is performed homomorphically on the encrypted model vectors.
In contrast, both MIFE and DMCFE take all encrypted model vectors as input and directly decrypt them to a single result.
Key generation and functional key derivation are excluded from the evaluation.
The comparison focuses solely on per-client ciphertext size, per-client encryption time, and combined decryption and aggregation time.
Ciphertext size is measured per client prior to aggregation.
All reported timings correspond to single measurements.

\Cref{fig:fe_vs_he_performance} reports the results for all metrics, with the two subfigures differing only in the use of a logarithmic scale on the x-axis.
The comparison indicates that, as the model parameter vector length increases, both MIFE and DMCFE incur substantially higher costs than HE across all considered metrics.

The model used in our FL experiments contains 83{,}126 parameters.
Based on interpolation from the measured data, at this model size MIFE produces approximately 1.9$\times$ larger ciphertexts than HE, requires over 700$\times$ longer per-client encryption time, and incurs more than 20$\times$ longer aggregation/decryption time.
Under the same conditions, DMCFE results in a similar increase in ciphertext size (approximately 1.9$\times$), more than 3{,}000$\times$ longer per-client encryption time, and more than 20$\times$ longer aggregation/decryption time relative to HE.
These measurements highlight the practical performance differences between HE and the FE schemes considered in this setting.

\section{Preliminaries}\label{sec:methods}
Our work integrates FedAvg with DP and S-HE.

\noindent \textbf{FedAvg.}
We consider the FedAvg framework~\cite{DBLP:conf/aistats/McMahanMRHA17}, where $N$ clients collaboratively train a global model $w^t$.
At round~$t$, the server sends $w^t$ to the clients.
Each client updates the model using its local dataset $d_i$, producing $w_i^{t+1}$. 
The server aggregates these local models by weighted averaging: $w^{t+1} {=} \sum_{i=1}\frac{|d_i|}{|\mathcal{D}|}w_i^{t+1}$, where $|D| = \sum_{i=1}^N |d_i|$.

\noindent \textbf{Selective Homomorphic Encryption~(S-HE).}
Unlike standard HE, which encrypts all model parameters, S-HE~\cite{DBLP:journals/corr/abs-2303-10837} selectively encrypts only sensitive parameters to reduce communication and computational overhead while preserving privacy. 
Sensitivity is defined as the magnitude of parameter gradients.
The server aggregates sensitivity scores across clients using weighted averaging and constructs a binary encryption mask $M$ by selecting the top $\eta$ fraction of parameters with the highest aggregated sensitivities.
Here, $M_m{=}1$ indicates that parameter $w_m$ is encrypted and $M_m{=}0$ otherwise.
The encryption ratio is defined as $\eta {=} \tfrac{1}{|w|}\sum_{m=1}^{|w|} M_m$, where $|w|$ is the total number of model parameters.
The resulting mask is shared with clients and remains fixed during training.

\noindent \textbf{Differential Privacy (DP).}
A mechanism $A$ satisfies $(\varepsilon, \delta)$-DP if, for any pair of adjacent datasets $d$ and $d'$ that differ in a single sample, and for any measurable output set $\mathcal{S}$, the following holds:
$$
\text{Pr}\left[A(d)\in \mathcal{S} \right] \leq e^\varepsilon\,\text{Pr}\left[A(d') \in \mathcal{S} \right] + \delta.
$$
We implement DP using DP-SGD~\cite{DBLP:conf/ccs/AbadiCGMMT016}, applying per-sample gradient clipping with maximum norm $C$ and additive Gaussian noise scaled by a noise multiplier $\sigma$.
To ensure a rigorous and fair comparison across different methods, our evaluation prioritizes empirical attack success rates as the measure of privacy leakage.
Consequently, we do not specify fixed theoretical privacy budgets $(\varepsilon, \delta)$. 
Instead, we adopt the DP-SGD mechanism~\cite{DBLP:conf/ccs/AbadiCGMMT016} and treat the noise multiplier $\sigma$ as the primary tunable parameter to control the DP defense strength.
We fix the clipping norm $C$ for each model and dataset to isolate the effect of varying noise strengths.

\noindent \textbf{Attacker Models.} 
\label{sec:threat_models}
The attacker aims to recover training data from client information exchanged during FL.
We study both active and passive data reconstruction attacks.
In both settings, the central server acts as the adversary attempting to recover private training data from clients, where the clients are \emph{honest}.
As the coordinator, the server knows the client data domains and model architecture.
Under FedAvg, clients transmit updated model parameters rather than explicit gradients.
However, the attacker can infer gradients by comparing successive plaintext parameter updates.
We assume that encrypted parameters are hidden and indistinguishable from random values.
If DP is used, the inferred gradients are further distorted in proportion to the noise magnitude.

Passive attacks, such as DLG and Inverting, follow an \emph{honest-but-cautious} model, in line with Hardy \etal~\cite{DBLP:journals/corr/abs-1711-10677}.
We assume that (i) the server faithfully executes the FL protocol, (ii) no collusion occurs, and (iii) the server attempts to infer as much information as possible from client updates.
We further assume that each gradient corresponds to a single image rather than a batch to allow tractable data reconstruction, as the reconstruction time increases substantially with batch size for the studied passive attacks, shown in \cite{DBLP:series/lncs/Zhu020}.

For active attacks, such as CAH and RTF, we follow the \emph{dishonest} model of Fowl \etal~\cite{DBLP:conf/iclr/FowlGCGG22}.
We retain all honest-but-curious assumptions but active attackers may additionally modify the model parameters or architecture.
Here, gradients correspond to batches of images, reflecting more realistic training conditions.

\section{Alternating Federated Learning (Alt-FL)}
\label{sec:rq1_integration}

In this section, we first provide an overview of the studied methods and then detail each proposed method.

\subsection{Overview}

We propose three interleaving-based methods (SI/DP, SI/HE, and PI) and compare them with the baseline MP.
The chosen method remains fixed throughout training.
SI/DP and SI/HE alternate between authentic and synthetic rounds using DP and S-HE, respectively.
PI alternates between DP and S-HE for privacy protection across training rounds.
In contrast, MP applies both DP and S-HE in every round.
These methods exhibit different privacy--quality--efficiency trade-offs.
We discuss how to select among them under our studied dataset and model in \Cref{sec:discussion}.

We remark that MP~\cite{DBLP:journals/corr/abs-2205-04330} is the most direct approach to combine DP and S-HE, leveraging their complementary strengths to reduce noise and encryption ratios while preserving privacy.
Our experiments show that although MP often preserves privacy, it is not always optimal.
This motivates the consideration of additional methods.

In our work, the MP baseline uses the encryption ratio $\eta$ as its configuration parameter, while PI, SI/DP, and SI/HE use the interleaving ratio defined in \Cref{sec:sidp}.

\subsection{Synthetic Interleaving with DP} \label{sec:sidp}

\begin{algorithm}[tbp]
\caption{\label{alg:si-dp}\smaller{Synthetic Interleaving using DP}}
\begin{smaller}
\SetKw{Where}{where}\SetKw{Upon}{upon}\SetKw{Receive}{receive}\SetKw{Is}{is}\SetKw{Send}{send}\SetKw{FromAll}{from all}\SetKw{To}{to}\SetKw{Broadcast}{broadcast}
\SetKwFunction{Enc}{Enc}\SetKwFunction{Dec}{Dec}
\SetKwData{IsAuth}{isAuth}\SetKwData{IsEnc}{isEnc}\SetKwData{True}{true}\SetKwData{False}{false}
\SetKwInOut{Input}{input}
\LinesNumbered

\Input{\begin{minipage}{\linewidth}
noise multiplier: $\sigma$ \\
clipping norm: $C$ \\
authentic dataset for client $i$: $d_{i, a}$\\
synthetic dataset for client $i$: $d_{i, s}$
\end{minipage}}
\BlankLine

\For{\textup{round} $t = 1,\,\dots,\,R$}{
    \IsAuth $\gets t \bmod \rho_{tot} < \rho_{tot}-\rho_{syn}$ \tcp*{where $\rho = \frac{\rho_{syn}}{\rho_{tot}}$ is the interleaving ratio} \label{ln:si-dp:2}
    \Upon client $i \in S$ \Receive model $w^{t-1}$ \Begin{ \label{ln:si-dp:3}
        \If{\IsAuth \Is \True}{ \label{ln:si-dp:4}
            $w^t_i \gets$ train $w^{t-1}$ using $d_{i,a}$ and DP-SGD with $ \sigma$ and $C$\; \label{ln:si-dp:5}
        }
        \lElse{$w^t_i \gets$ train $w^{t-1}$ using $d_{i,s}$} \label{ln:si-dp:6}
        \Send $w_i^t$ \To server\; \label{ln:si-dp:7}
    }

    \Upon server \Receive $w_i^t$ \FromAll client $i \in S$ \Begin{ \label{ln:si-dp:8}
        $W^t \gets \{w^t_1,\,\dots,\,w^t_{|S|}\}$\tcp*{the set of all received client models} \label{ln:si-dp:9}
        $w^t \gets$ aggregate $W^t$\; \label{ln:si-dp:10}
        \Broadcast $w^t$ \To all clients\; \label{ln:si-dp:11}
    }
}

\end{smaller}
\end{algorithm}

We first explain the notion of interleaving.
A training round is \emph{authentic} if it uses authentic data and \emph{synthetic} otherwise.
We formalize the interleaving schedule via the synthetic interleaving ratio
$\rho := \rho_{syn}/\rho_{tot}$,
where  $\rho_{syn} {\geq 0}$ is the number of synthetic rounds, and $\rho_{tot} {>} 0 $ is the total number of rounds, satisfying $\rho_{tot} {\geq} \rho_{syn}$.
In practice, we express this ratio in the lowest possible integer terms; \eg 
$\rho {=} 0.25$ corresponds to $\rho_{syn} {=} 1$ and $\rho_{tot} {=} 4$, meaning that one of four rounds is synthetic.

The SI/DP method, described in \Cref{alg:si-dp}, interleaves synthetic and authentic data while using DP as the privacy mechanism in authentic rounds.
Therefore, the algorithm extends the standard FL workflow with authentic and synthetic rounds. 
At \Cref{ln:si-dp:2}, the algorithm determines whether the current round is authentic by evaluating: $t\; {\text{mod}}\; \rho_{tot} < \rho_{tot} {-} \rho_{syn}$.

The client-side operations are given in \Crefrange{ln:si-dp:3}{ln:si-dp:7}.
Each client trains the received model according to the round type, using DP-SGD in authentic rounds and training on synthetic data without DP in synthetic rounds.

The server-side behavior is described in \Crefrange{ln:si-dp:8}{ln:si-dp:11}.
The server aggregates client updates and broadcasts the resulting model without distinguishing between round types.
We remark that our methods require no server-side operations beyond model aggregation, \eg no server-side DP, as the server is assumed adversarial and untrusted.

\subsection{Synthetic Interleaving with HE} \label{sec:sihe}

\begin{algorithm}[tbp]
\caption{\label{alg:si-he}\smaller{Synthetic Interleaving using Selective HE}}
\begin{smaller}
\SetKw{Where}{where}\SetKw{Upon}{upon}\SetKw{Receive}{receive}\SetKw{Is}{is}\SetKw{Send}{send}\SetKw{FromAll}{from all}\SetKw{To}{to}\SetKw{Broadcast}{broadcast}
\SetKwFunction{Enc}{Enc}\SetKwFunction{Dec}{Dec}
\SetKwData{IsAuth}{isAuth}\SetKwData{IsEnc}{isEnc}\SetKwData{True}{true}\SetKwData{False}{false}
\SetKwInOut{Input}{input}
\LinesNumbered

\Input{\begin{minipage}{\linewidth}
\fbox{\begin{minipage}{24mm}
encryption mask: $M$
\end{minipage}} \\
authentic dataset for client $i$: $d_{i, a}$\\
synthetic dataset for client $i$: $d_{i, s}$
\end{minipage}}
\BlankLine
\For{\textup{round} $t = 1,\,\dots,\,R$}{\label{ln:si-he:1}
    \IsAuth $\gets t \bmod \rho_{tot} < \rho_{tot}-\rho_{syn}$\tcp*{where $\rho = \frac{\rho_{syn}}{\rho_{tot}}$ is the interleaving ratio}\label{ln:si-he:2}
    \Upon client $i \in S$ \Receive model $w^{t-1}$ \Begin{\label{ln:si-he:3}
        \fbox{\begin{minipage}{56mm}
        \IsEnc $\gets$ test if $w^{t-1}$ is encrypted\;
        \lIf{\IsEnc \Is \True}{$w^{t-1} \gets$ \Dec{$w^{t-1}$, $M$}}
        \end{minipage}}\label{ln:si-he:4}\\
        \If{\IsAuth \Is \True}{\label{ln:si-he:5}
            \fbox{\begin{minipage}{36mm}
            $w^t_i \gets$ train $w^{t-1}$ using $d_{i,a}$\;
            $w^t_i \gets$ \Enc{$w^t_i$, $M$}\;
            \end{minipage}}\label{ln:si-he:6}
        }
        \lElse{$w^t_i \gets$ train $w^{t-1}$ using $d_{i,s}$}\label{ln:si-he:7}
        \Send $w_i^t$ \To server\;\label{ln:si-he:8}
    }

    \Upon server \Receive $w_i^t$ \FromAll client $i \in S$ \Begin{\label{ln:si-he:9}
        $W^t \gets \{w^t_1,\,\dots,\,w^t_{|S|}\}$\tcp*{the set of all received client models}\label{ln:si-he:10}
        \fbox{\begin{minipage}{60mm}
        \If{\IsAuth \Is \True}{
            $w^t \gets$ HE aggregation of $W^t$ using mask $M$\;
        }
        \lElse{$w^t \gets$ aggregate $W^t$}
        \end{minipage}}\label{ln:si-he:11}\\
        \Broadcast $w^t$ \To all clients\;\label{ln:si-he:12}
    }
}

\end{smaller}
\end{algorithm}

We describe SI/HE in \Cref{alg:si-he} and use \fbox{boxed} lines to highlight differences from~\Cref{alg:si-dp}.
SI/HE follows the same structure as SI/DP, but replaces DP with S-HE.

In SI/HE, clients train on plaintext models, decrypting the received global model when necessary (\Cref{ln:si-he:4}).
During authentic rounds, clients encrypt updates using the mask $M$ before transmission.
The  server then performs homomorphic aggregation over the encrypted updates, also using $M$, and broadcasts the results.

\subsection{Privacy Interleaving} \label{sec:pi}

\begin{algorithm}[tbp]
\caption{\label{alg:pi}\smaller{Privacy Interleaving}}
\begin{smaller}
\SetKw{Where}{where}\SetKw{Upon}{upon}\SetKw{Receive}{receive}\SetKw{Is}{is}\SetKw{Send}{send}\SetKw{FromAll}{from all}\SetKw{To}{to}\SetKw{Broadcast}{broadcast}
\SetKwFunction{Enc}{Enc}\SetKwFunction{Dec}{Dec}
\SetKwData{IsAuth}{useHE}\SetKwData{IsEnc}{isEnc}\SetKwData{True}{true}\SetKwData{False}{false}
\SetKwInOut{Input}{input}
\LinesNumbered

\Input{\fbox{\begin{minipage}{40mm}
    encryption mask: $M$\\
    noise multiplier: $\sigma$\\
    clipping norm: $C$\\
    authentic dataset for client $i$: $d_{i, a}$
\end{minipage}}}
\BlankLine
\For{\textup{round} $t = 1,\,\dots,\,R$}{
    \fbox{\IsAuth} $\gets t \bmod \rho_{tot} < \rho_{tot}-\rho_{syn}$ \tcp*{where $\rho = \frac{\rho_{syn}}{\rho_{tot}}$ is the interleaving ratio} 
    \Upon client $i \in S$ \Receive model $w^{t-1}$ \Begin{ \label{ln:pi:3}
        \fbox{\begin{minipage}{56mm}
        \IsEnc $\gets$ test if $w^{t-1}$ is encrypted\;
        \lIf{\IsEnc \Is \True}{$w^{t-1} \gets$ \Dec{$w^{t-1}$, $M$}}
        \end{minipage}}\\
        \If{\IsAuth \Is \True}{
            \fbox{\begin{minipage}{36mm}
            $w^t_i \gets$ train $w^{t-1}$ using $d_{i,a}$\;
            $w^t_i \gets$ \Enc{$w^t_i$, $M$}\;
            \end{minipage}}
        }
        \fbox{\begin{minipage}{56mm}
        \Else{
            $w^t_i \gets$ train $w^{t-1}$ using $d_{i,a}$ and DP-SGD with $\sigma$ and $C$\;
        }
        \end{minipage}}\\ \label{ln:pi:7}
        \Send $w_i^t$ \To server\; \label{ln:pi:8}
    }

    \Upon server \Receive $w_i^t$ \FromAll client $i \in S$ \Begin{
         $W^t \gets \{w^t_1,\,\dots,\,w^t_{|S|}\}$\tcp*{the set of all received client models}
        \fbox{\begin{minipage}{60mm}
        \If{\IsAuth \Is \True}{
            $w^t \gets$ HE aggregation of $W^t$ using mask $M$\;
        }
        \lElse{$w^t \gets$ aggregate $W^t$}
        \end{minipage}}\\
        \Broadcast $w^t$ \To all clients\;
    }
}

\end{smaller}
\end{algorithm}

Privacy-preserving synthetic data may incur high generation costs.
To reduce this overhead without relying on synthetic data, we introduce PI.

PI alternates between DP and S-HE according to the interleaving ratio $\rho$, using only authentic data.
A round is denoted \emph{DP round} when DP is applied and \emph{HE round} when S-HE is used.
The algorithm for PI is shown in \Cref{alg:pi}, with differences from \Cref{alg:si-dp} highlighted using \fbox{boxed} lines.

The variable indicating the round type is renamed to \texttt{useHE}, reflecting whether S-HE should be applied.
On the client side (\Crefrange{ln:pi:3}{ln:pi:8}), S-HE is used when \texttt{useHE} is true, and DP otherwise.
The server-side behavior follows \Cref{alg:si-he}, applying the HE aggregation when necessary.

\section{Privacy Protection Levels}
\label{sec:privacy_levels}
We define privacy protection levels by selecting method-specific privacy parameters.
As mentioned, SI/DP uses only DP, SI/HE only S-HE, while PI and MP combine DP and HE.
Thus, parameter selection differs across~methods:

\noindent $\bullet$ SI/DP selects the smallest noise multiplier $\sigma$, achieving a near-zero (${<}0.5\%$) attack success rate.

\noindent $\bullet$ SI/HE selects the smallest encryption ratio $\eta$, achieving a near-zero attack success rate;

\noindent $\bullet$ PI selects the smallest noise multiplier $\sigma$ and encryption ratio $\eta$ that individually achieve a near-zero success rate;

\noindent $\bullet$
MP fixes $\eta{\in}\{0,0.2,0.4,0.6,0.8,1\}$, and, for each $\eta$, selects the smallest $\sigma$, achieving a near-zero success rate when DP and S-HE are used simultaneously.
Thus, $\sigma$ varies with $\eta$.

\begin{figure*}[htbp]
    \centering
    \includegraphics[width=\linewidth]{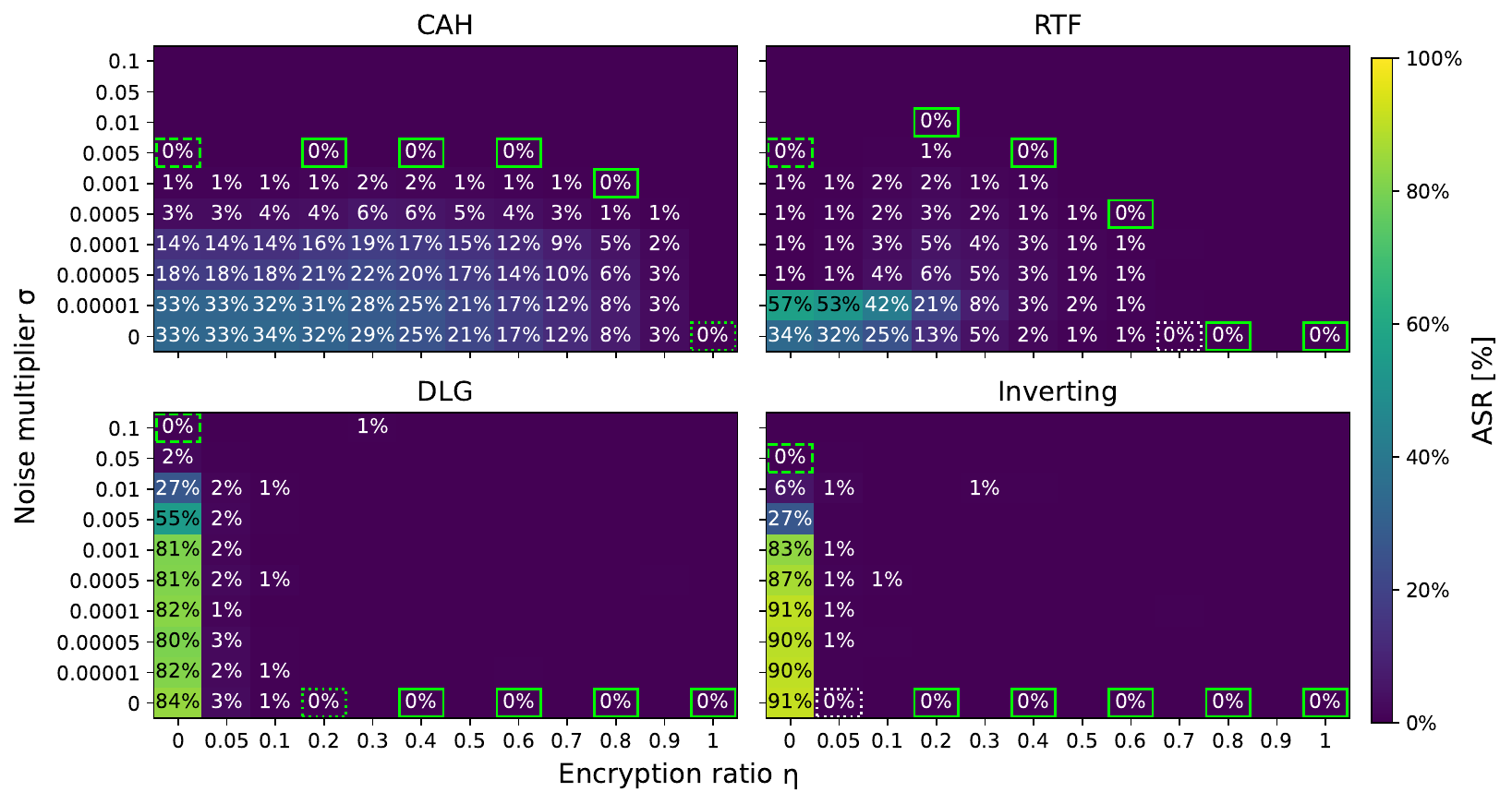}
    \caption{Attack success rates (ASR) on model protected by DP (varying noise) and/or S-HE (varying encryption ratio) for each studied attack. Success rates are rounded to the nearest percent, and 0\% success rates are omitted for clarity.}
    \label{fig:rq2_succ_rates}
\end{figure*}

\smallskip
To define privacy levels, we evaluate each attack on models protected by DP, HE, or both, varying $\sigma {\in} [0, 0.1]$ and $\eta {\in} [0, 1]$.
Each attack is repeated 1000 times to obtain the success rate matrices in \Cref{fig:rq2_succ_rates}.
From the matrices, we extract relevant privacy parameters as follows:

\noindent $\bullet$ \textbf{SI/DP}:  $\sigma$ is selected from the \emph{left-most column} at the first cell, achieving a near-zero (${<}0.5\%$) success rate, scanning from bottom to top, highlighted with a \emph{dashed} outline;

\noindent $\bullet$ \textbf{SI/HE}: $\eta$ is selected from the \emph{bottom-most row} at the first cell, achieving a near-zero success rate, scanning from left to right, highlighted in a box with a \emph{dotted} outline;

\noindent $\bullet$ \textbf{PI}: $\sigma$ and $\eta$ are selected independently using the same procedures as SI/DP and SI/HE, respectively;

\noindent $\bullet$ \textbf{MP}: For each fixed $\eta{\in}\{0, 0.2, 0.4, 0.6, 0.8, 1\}$, $\sigma$ is selected from the corresponding column in the first cell, achieving a near-zero success rate, scanning from bottom to top, highlighted with \emph{green} lines.

\smallskip

\noindent \textbf{Defining Privacy Levels.}
Using the matrices in \Cref{fig:rq2_succ_rates}, we define \emph{attack-specific} privacy levels corresponding to the minimum privacy parameters required to defend against each specific attack (DLG, Inverting, CAH, and RTF). 
We further define \emph{extended} privacy levels, denoted \emph{Infimum} and \emph{Supremum}.
The Infimum level is the greatest lower bound of the attacker-specific levels, and the Supremum level is the least upper bound.

We also consider privacy levels commonly used in DP-SGD evaluations and FL tutorials~\cite{Ahmed2024DPFL_COVID19,tensorflow_federated_dp_tutorial}, referred to as the \emph{standard} privacy levels.
The standard privacy levels are specified with $\eta{=}1$ and $\sigma  {\in} \{0.1, 0.25, 0.5, 0.75, 1\}$, spanning noise multipliers more commonly used in real-world.
We denote these standard levels as \emph{Std-X}, where \emph{X} corresponds to the $\sigma$ value.
We exclude the SI/HE method from evaluations on standard levels, as it does not use DP.

The MP method constitutes a special case in which the encryption ratio $\eta$ is fixed by the MP configuration and remains independent of the privacy level, whereas the noise multiplier is determined by the selected level.

\begin{table}[tbp]
\caption{DP noise multipliers $\sigma$ selected for each encryption ratio $\eta$ for all privacy levels (PL)}
\label{tab:privacy_level_params}
\centering
\begin{tblr}{
  width=\columnwidth,
  colspec = {l|llllllll},
  colsep = 3pt,
  rowsep = 0.5pt,
  row{1} = {font=\bfseries},
  column{1} = {font=\bfseries},
}
\toprule
\diagbox{PL}{$\eta$} & 0 & 0.05 & 0.2 &  0.4 & 0.6 & 0.7 & 0.8 & 1 \\
\midrule

Std-1.0 & \SetCell[c=7]{c} 1.0 &  &  &   &  &  &  &  \\
Std-0.75 & \SetCell[c=7]{c} 0.75 &  &  &  &  &  &  &  \\
Std-0.5 & \SetCell[c=7]{c} 0.5 &  &  &  &  &  &  &  \\
Std-0.25 & \SetCell[c=7]{c} 0.25 &  &  &  &  &  &  &  \\
Std-0.1 & \SetCell[c=7]{c} 0.1 &  &  &  &  &  &  &  \\
\midrule
Supremum & 0.1 & 0.05 & 0.01 & 0.005 & 0.005 & 0.005 & 0.001 & \SetCell{cmd=\fbox} 0 \\
DLG & 0.1 & 0.05 & \SetCell{cmd=\fbox} 0 & 0 & 0 & 0 & 0 & 0 \\
Inverting & 0.05 & \SetCell{cmd=\fbox} 0 & 0 & 0 & 0 & 0 & 0 & 0 \\
RTF & 0.005 & 0.005 & 0.01  & 0.005 & 0.0005 & \SetCell{cmd=\fbox} 0 & 0 & 0 \\
CAH & 0.005 & 0.005 & 0.005 & 0.005  & 0.005 & 0.005 & 0.001  & \SetCell{cmd=\fbox} 0 \\
Infimum &  0.005 & \SetCell{cmd=\fbox} 0 & 0 & 0 & 0 & 0 & 0 & 0 \\
\bottomrule

\end{tblr}
\end{table}

\Cref{tab:privacy_level_params} reports the selected $\sigma$ values for each privacy level and encryption ratios $\eta$ (columns).
Unused $\eta$ values are omitted, and the minimum $\eta$ for each privacy level is indicated by a boxed \fbox{0}.
For standard privacy levels, their $\sigma$ values remain fixed regardless of $\eta$.

These results in \Cref{tab:privacy_level_params} show that the studied attacks can be mitigated with relatively low noise multipliers ($\sigma {\le} 0.1$), which correspond to higher privacy budgets $\varepsilon$ and thus weaker DP guarantees.
Although PI and MP combine DP and S-HE, there is no formal framework for composing these privacy protection mechanisms into a single quantifiable guarantee.
We therefore assess privacy guarantee by empirically evaluating the DP component alone and report the resulting $\varepsilon$ values induced by the considered noise multipliers in \Cref{sec:noise_mult_to_budget}.

\section{Correlating Noise Multipliers to DP Budgets}
\label{sec:noise_mult_to_budget}

\begin{table}[tbp]
\caption{Average cumulative privacy budget $\varepsilon$ at convergence for noise multipliers $\sigma \ge 0.05$}
\label{tab:dp_eps}
\centering
\begin{tblr}{
  width=\textwidth,
  colspec = {l|llllll},
  row{1} = {font=\bfseries},
  column{1} = {font=\bfseries},
  row{1,8-Z} = {bg=white},
}
\toprule
$\sigma$ & 0.05 & 0.1 & 0.25 &  0.5 & 0.75 & 1 \\

\midrule

Mantissa & 1.17 & 1.47 & 1.09 & 5.36 & 1.25 & 6.63 \\
Exponent & $10^5$ & $10^4$ & $10^3$ &  $10^1$ & $10^1$ & $10^0$  \\

\bottomrule
\end{tblr}
\end{table}

\begin{figure}[htbp]
    \centering
    \includegraphics[width=\linewidth]{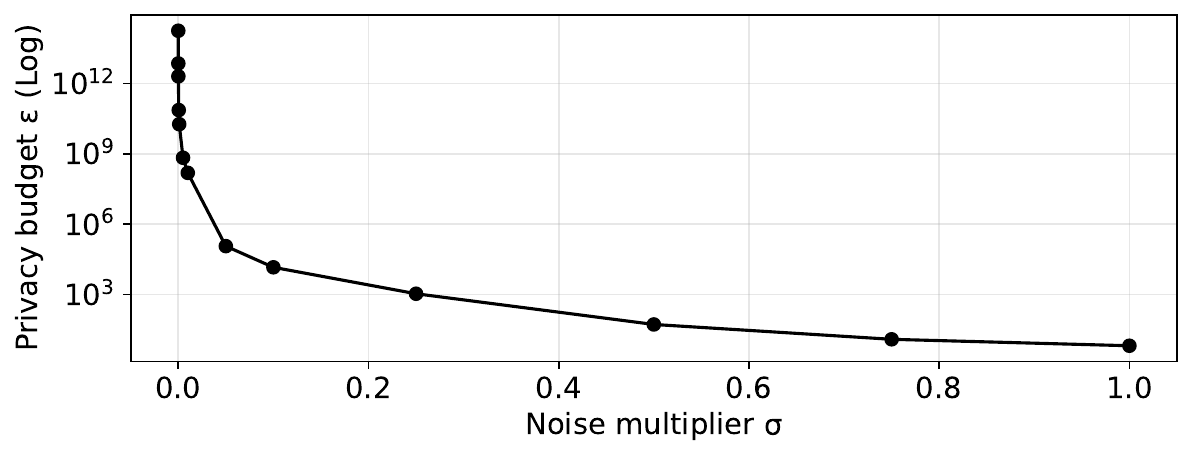}
    \caption{Average cumulative privacy budget $\varepsilon$ at convergence as a function of the noise multiplier $\sigma$.}
    \label{fig:dp_eps}
\end{figure}

To provide more context for the noise multipliers used in our experiments for DP, we report the cumulative privacy budget $\varepsilon$ associated with different values of the noise multiplier $\sigma$.
These results are intended to serve as a reference for interpreting the scale of privacy guarantees induced by the studied noise levels.

The privacy budget is computed using the R\'enyi Differential Privacy (RDP) accountant~\cite{DBLP:conf/csfw/Mironov17}.
For all experiments, we fix the clipping norm to $C {=} 4.7$ and set $\delta {=} 1/N$, where $N$ denotes the dataset size.
Although we do not formally define a global \((\varepsilon,\delta)\)-DP guarantee, the reported $\varepsilon$ values correspond to this fixed choice of $\delta$.

For each noise multiplier  $\sigma {\in} [0.00005, 1]$, we train the model until convergence, as determined by the early stopping criterion defined in earlier sections.
Because convergence occurs at different training rounds for different noise levels, the resulting cumulative privacy budget varies accordingly.
Each experiment is repeated 10 times.
In the federated setting, we compute $\varepsilon$ independently for each client and report the maximum $\varepsilon$ across clients, averaged over the 10 runs.

\Cref{fig:dp_eps} shows the average cumulative privacy budget $\varepsilon$ at convergence as a function of the noise multiplier $\sigma$.
The raw numerical values for $\sigma \ge 0.05$ are reported in \Cref{tab:dp_eps}, using a mantissa--exponent format to emphasize the scale of the resulting privacy budgets.
Results for $\sigma < 0.05$ are omitted, as they yield cumulative privacy budgets exceeding $10^{8}$.

\section{Experimental Settings} \label{sec:evaluation}

We conduct two types of experiments: FL training to evaluate learning quality and efficiency, and empirical data reconstruction attacks to evaluate the resilience of the methods against such attacks.

\Cref{app:glossary} provides a glossary of abbreviations and a table of mathematical notation used throughout this work.

\noindent\textbf{Configuration Parameters.}
Each method is configured by a method-specific parameter.
MP is configured by the encryption ratio $\eta {\in} \{0, 0.2, 0.4, 0.6, 0.8, 1\}$.
The interleaving methods are configured by the interleaving ratio $\rho {\in} \{0, \tfrac{1}{4}, \tfrac{2}{5}, \tfrac{1}{2}, \tfrac{3}{5}, \tfrac{3}{4}\}$, and PI additionally includes $\rho{=}1$.
Here, $\rho{=}0$ denotes HE-only and $\rho{=}1$ DP-only, \ie exclusive use of the respective privacy protection mechanism.

\noindent\textbf{Data Distribution and Augmentation.}
We evaluate various non-IID settings, simulating non-IID authentic data via a per-class Dirichlet distribution with $\alpha {\in} \{0.25, 0.5, 1\}$, where smaller $\alpha$ indicates greater non-IIDness.
Synthetic data are always IID, as we assume clients can generate them locally.

We also evaluate the effect of synthetic data augmentation.
We define the augmentation ratio $r$ as the fraction of synthetic data in local training and consider $r {\in} \{0, 0.25, 0.5\}$.
Results for $r{=}0.5$ are omitted due to inferior quality--cost trade-offs compared to $r{=}0.25$.
When augmentation is enabled, $\alpha$ is fixed at 0.5 to isolate the effects of augmentation.

Synthetic data are generated using a differentially private diffusion model following Dockhorn \etal~\cite{DBLP:journals/tmlr/DockhornCVK23} with $\epsilon{=}10$, producing a synthetic dataset matching the original size.
The diffusion model is trained once per dataset, and the resulting data are assumed to be privacy preserving.

\noindent\textbf{FL Training.}
We use LeNet-5~\cite{DBLP:journals/pieee/LeCunBBH98} on CIFAR-10~\cite{krizhevsky2009learning} with three clients, following~\cite{DBLP:journals/corr/abs-2303-10837}.
FL training is simulated on a single machine with Dirichlet-partitioned data. 

We evaluate all combinations of privacy levels, methods, method-specific configurations, and either non-IIDness parameter $\alpha$ or augmentation ratio $r$.
All other hyperparameters (\eg optimizer, learning rate, batch size, and local epochs) are fixed across configurations.

Training is performed on an Intel Xeon Platinum 8358 CPU with four allocated cores.
For each configuration, we run training 10 times and report the mean after discarding the best and worst values.
We apply a moving average to accuracy and declare convergence when the improvement remains below $0.1\%$ for 10 consecutive rounds.
The moving average window size is set to 10 for MP.
For interleaving-based methods, the size is set to 8 when $\rho {\in} \{\tfrac{1}{4}, \tfrac{3}{4}\}$ and 10 otherwise.
These window sizes achieve improved convergence times in our empirical tests.

We measure learning quality in terms of accuracy and efficiency, as reflected in system costs.
Thus, we evaluate accuracy, convergence time, communication cost, and computational cost.
Accuracy is reported as the best model accuracy achieved on a held-out test set during training.
Convergence time is the average number of rounds until convergence.
Communication cost is measured per client and includes the total size of transmitted models and ciphertexts until convergence.
Computational cost is the total time until convergence, including cryptographic operations.

\noindent\textbf{Privacy Attacks on FL Models.} 
To quantify privacy protection, we measure the success rates of privacy attacks against the trained FL models.
We perform attacks exhaustively on all training configurations, excluding those with synthetic augmentation, as augmentation is not intended to serve as a privacy protection mechanism in this work.

For each configuration, we perform 200 attack trials and average the results.
For PI, we apply DP in $200 {\cdot} \rho$ trials and HE in the remaining $200 {\cdot} (1 {-} \rho)$ trials.
For SI/DP and SI/HE, attacks are performed on authentic rounds 200 times.
Note that our considered reconstruction attacks are simulated independently and can be carried out on arbitrary training rounds.
Therefore, they are not restricted to models obtained from the training~experiments.

We quantify attack success using the \emph{Image Identifiability Precision} (IIP) metric~\cite{DBLP:conf/cvpr/YinMVAKM21}.
IIP measures the fraction of reconstructed images that can be correctly matched to authentic images via nearest-neighbor search.
We report the maximum IIP score across attacks, representing the worst-case privacy risk.

\section{Privacy Preservation}
\label{sec:privacy_preservation}

\begin{figure}[tbp]
    \centering
    \includegraphics[width=1\linewidth]{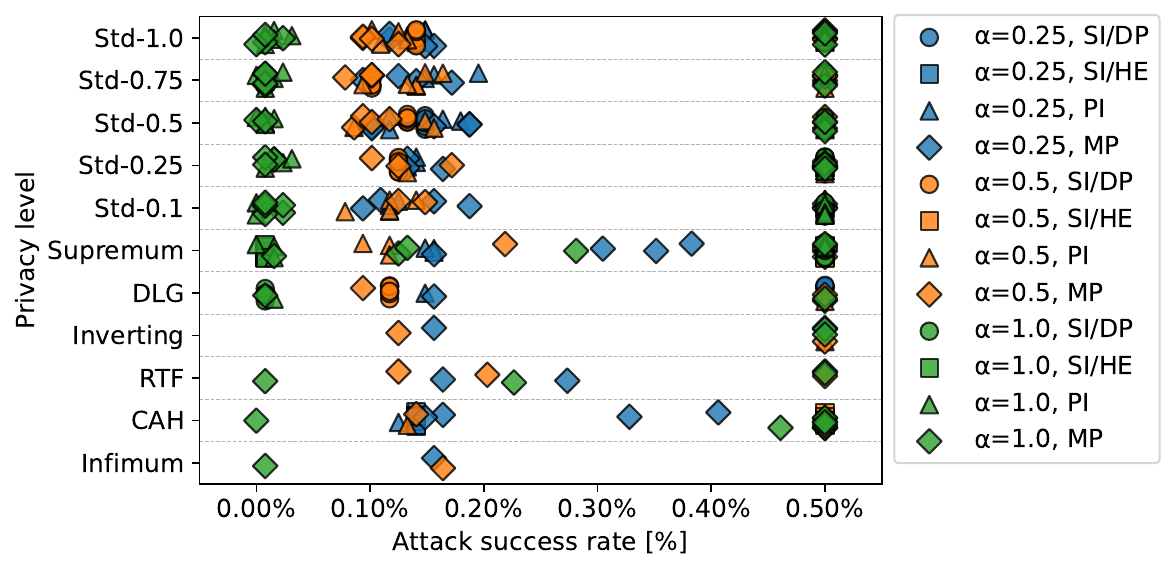}
    \caption{All methods and configurations with a success rate of 0.5\% or lower are shown. The legend on the right lists the corresponding methods and levels of non-IIDness. Each symbol appears once for every configuration whose success rate does not exceed 0.5\%.}
    \label{fig:succ_rate_0_5_per}
\end{figure}

To answer RQ1, we identify methods achieving limited attack success rates at each privacy level.
\Cref{fig:succ_rate_0_5_per} shows method configurations with success rates ${\le}0.5\%$.
We observe that non-IIDness affects success rates, as indicated by variations between $0\%$ and $0.2\%$, which necessitate further analysis to determine suitable configurations.

\noindent \textbf{Threshold for Privacy Protection.}
We define a threshold $t_{asr}$ to determine suitable attack success rates.
A configuration is considered suitable if its success rate does not exceed the minimum observed success rate (for the same privacy level and $\alpha$) plus $t_{asr}$.
As lower success rates indicate stronger privacy protection, this criterion systematically selects suitable method configurations.

The threshold is intended to enable interpretable method selection.
However, overly restrictive thresholds can exclude suitable candidates, while large thresholds may admit an excessive number of configurations.
We therefore vary $t_{asr}{\in}[0.001, 0.01]$ in increments of 0.001 and stop when further increases do not significantly change the selected methods or when it selects an excessive number of configurations.
We select $t_{asr} {=} 0.005$ (0.5\%), beyond which the selections do not change significantly.

\begin{table*}[tbp]
\caption{Methods and their configurations that provide the best defenses against the privacy attacks under varying data distribution, with $t_{asr}=0.5\%$. Note that the case with synthetic augmentation is identical to $\alpha {=} 0.5$. Methods marked with $*$  are  those in which the chosen configurations vary by at most two configurations across different privacy levels}
\label{tab:best_privacy_collapsed}
\centering
\begin{tblr}{
  width = \textwidth,
  rowsep = 0.5pt,
  colspec = {X[0.4,l] X[1,l] X[1,l] X[1,l]},
  row{1} = {font=\bfseries},
  hlines, vlines
}
 & {$\alpha = 0.25$} & {$\alpha = 0.5$} & {$\alpha = 1$} \\

Std-1.0 to Std-0.25 &
{SI/DP, PI\textsuperscript{*}, MP\textsuperscript{*}} &
{SI/DP, PI\textsuperscript{*}, MP\textsuperscript{*}} &
{SI/DP, PI\textsuperscript{*}, MP\textsuperscript{*}} \\

Std-0.1 &
{PI,\\ MP: $\eta \in \{0.2, 0.4, 0.6, 1\}$}&
{SI/DP,\\ PI (excl. $\rho = 1$),\\ MP (excl. $\eta=1$)} &
{SI/DP, PI, MP} \\

Supremum &
{SI/HE, PI, MP} &
{SI/DP, SI/HE,\\ PI (excl. $\rho\in\{0, \tfrac{3}{5}\}$), \\ MP: $\eta \in \{0.2, 0.6, 1\}$} & 
{SI/DP, SI/HE, PI,\\ MP (excl. $\eta=0$)} \\

DLG &
{SI/DP,\\ DP (SI/DP, PI, MP), HE (MP)} &
{SI/DP,\\ DP (SI/DP, PI, MP), HE (MP)} &
{SI/DP,\\ DP (SI/DP, PI, MP), HE (MP)} \\

Inverting &
{DP (MP), HE (MP)} &
{DP (PI, MP), HE (MP)} &
{DP (MP), HE (MP)} \\

RTF &
{MP: $\eta\in\{0.2, 0.4, 1\}$ \\ HE  (MP)} &
{MP: $\eta\in\{0.2, 0.4, 1\}$ \\ HE  (MP)} &
{MP: $\eta\in\{0.2, 0.4, 1\}$ \\ HE  (MP)} \\

CAH &
{SI/HE, \\ MP (excl. $\eta=0$), \\ HE (MP, PI, SI/HE) } &
{SI/HE, \\ MP (excl. $\eta=0$), \\ HE (MP, PI, SI/HE) } &
{SI/HE, \\ MP (excl. $\eta=0$), \\ HE (MP, PI, SI/HE) } \\

Infimum &
{HE (MP)} &
{HE (MP)} &
{HE (MP)} \\
\end{tblr}

\end{table*}

\noindent \textbf{Methods for Privacy Protection.}
\Cref{tab:best_privacy_collapsed} lists all method configurations that meet $t_{asr}$, highlighting those that effectively defend against the studied attacks at each privacy level.
For clarity, we mark configurations equivalent to DP-only and HE-only as DP and HE, respectively.
We list only the method name when all configurations meet the threshold.
The augmented case at $r {=} 0.5$ is omitted, as it matches the non-augmented results. 

At standard privacy levels (Std-1.0 to Std-0.1), most methods perform well.
However, at Std-0.1, when the data is less IID ($\alpha{=}0.25$), SI/DP fails to defend but succeeds when the data becomes more IID.
For attack-specific and extended privacy levels, the set of methods that can effectively defend becomes more nuanced.
At the Supremum level, all methods provide suitable defense except SI/DP at $\alpha{=}0.25$, mirroring Std-0.1.
For DLG and Inverting, only configurations with DP-only or full encryption are selected, as these levels have high noise multipliers but low encryption ratios.
At the CAH level, only SI/HE and MP provide adequate defense due to a weak noise multiplier ($\sigma{=}0.005$).
Moreover, the configurations selected for the RTF privacy level are largely a subset of CAH's, since the privacy parameters of RTF are equal to or lower than those of CAH.
At the Infimum level, only full encryption can provide adequate defense.

MP consistently provides suitable defense across all privacy levels. 
However, as shown later, MP exhibits noticeable drawbacks in  learning quality and efficiency.
This strong privacy protection can be explained by two factors.
First, MP combines HE and DP, whose effects complement each other, thereby strengthening resistance to the studied attacks.
Second, MP uses the encryption ratio $\eta$ as the parameter that controls the selective HE ratio, independent of the $\eta$ defined by the privacy level. 
Consequently, MP with $\eta {=} 1$ defends against all studied attacks.
This holds for the DLG, Inverting, and Infimum levels, which use weaker encryption ratios that fail to defend against all studied attacks.

\section{System Costs vs. Learning Quality}
\label{sec:system_costs_vs_learning_quality}

To address RQ2, we examine the accuracy and system costs under privacy preservation and identify method configurations that achieve the best trade-offs.

\noindent \textbf{Thresholds for Accuracy and System Costs.}
We first consider configurations that satisfy the threshold $t_{asr}$ defined in \Cref{sec:privacy_preservation}, representing privacy-preserving method configurations.
To broaden exploration, we relax the threshold to $t_{asr} {=} 0.01$.

\begin{figure}[tbp]
    \centering
    \includegraphics[width=\linewidth]{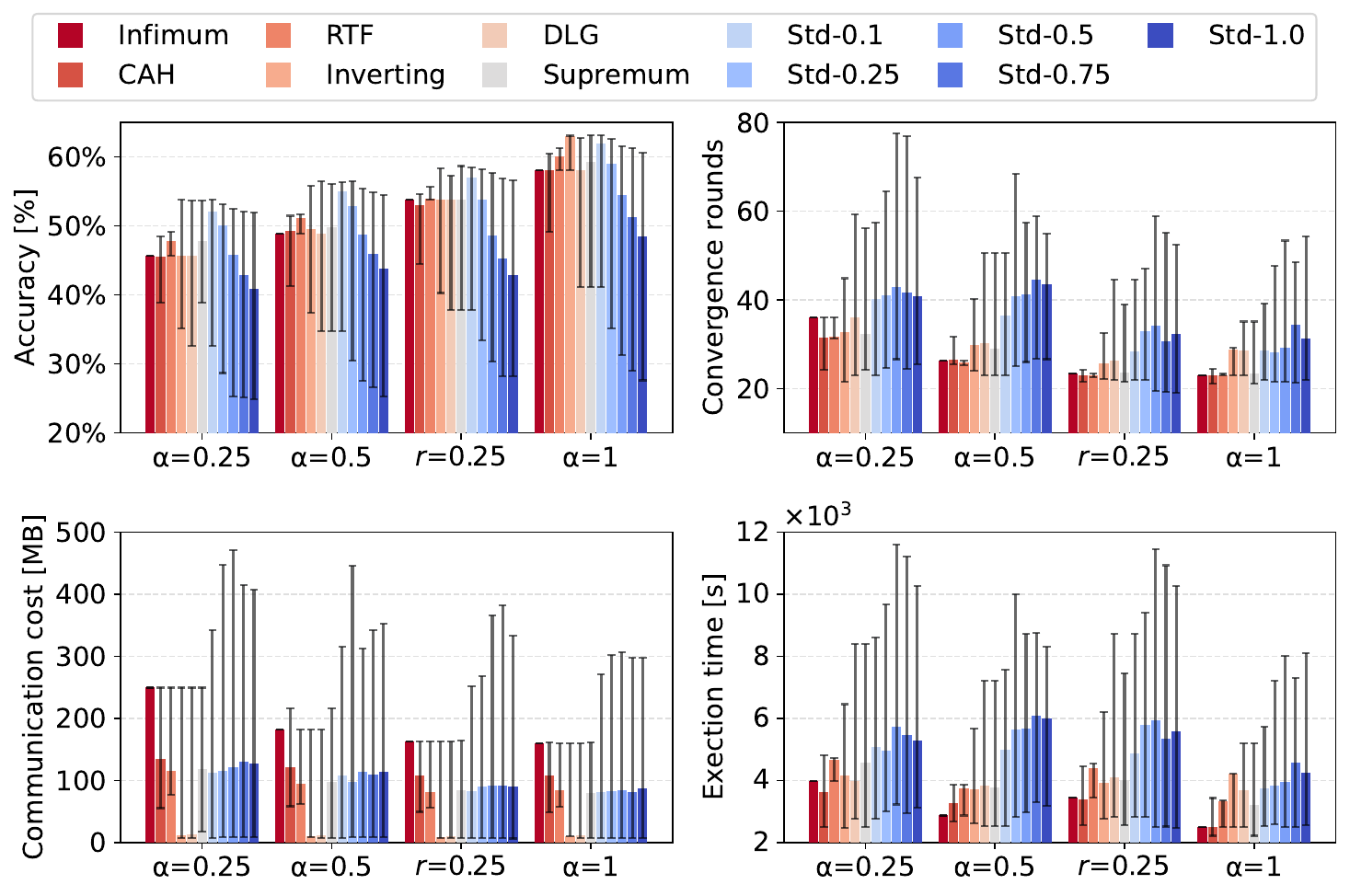}
    \caption{Bar charts of accuracies and system costs for the methods and configurations that meet the threshold $t_{asr} {=} 0.01$.
    For each privacy level and non-IIDness parameter $\alpha$, black error bars indicate the best and worst metrics, while colored bars represent the median values. The $r {=} 0.25$ corresponds to the synthetic data augmentation case.
    }
    \label{fig:RQ4_acc_costs}
\end{figure}

\Cref{fig:RQ4_acc_costs} reports the best, median, and worst metric values for configurations meeting $t_{asr}$.
The wide error intervals indicate substantial variations in performance despite satisfying $t_{asr}$, motivating additional thresholds $t_{acc}$ and $t_{sc}$ to balance accuracy and system costs.

To identify cost-efficient configurations, we apply $t_{sc}$ sequentially to each system cost metric (in the order of communication cost, computational cost, and convergence time).
A configuration satisfies the system cost threshold if its cost is within $t_{sc}\times 100\%$ of the observed minimum cost for the same privacy level and non-IIDness.
We vary $t_{sc}$ in $[0.1, 0.5]$ in increments of 0.1.
We set the final threshold to $t_{sc}{=}0.5$, as smaller values select too few configurations.

To identify configurations that maintain high accuracy \emph{and} reasonable resource utilization while preserving privacy, we also introduce the accuracy threshold $t_{acc}$.
A configuration meets the accuracy threshold if its accuracy is at least the observed maximum accuracy (for the same privacy level and non-IIDness) minus $t_{acc}$.
To determine the value for $t_{acc}$, we fix $t_{asr}{=}0.01$ and $t_{sc}{=}0.5$, then vary $t_{acc}$ over $[0.01, 0.1]$ in increments of $0.01$.
We set $t_{acc} {=} 0.04$, beyond which the selections do not change significantly while accuracy degrades.
The thresholds are applied in the following order: $t_{asr}$, $t_{acc}$, and $t_{sc}$, prioritizing accuracy over system cost.

To answer RQ2, we consider methods and configurations selected using thresholds $t_{asr} {=} 0.01$, $t_{acc} {=} 0.04$, and $t_{sc} {=} 0.5$.
We apply the thresholds sequentially: first $t_{asr}$, then $t_{acc}$, and finally $t_{sc}$.
The resulting configurations represent those that best balance accuracy and system costs while maintaining privacy.

\begin{figure}[t]
    \centering
    \includegraphics[width=\linewidth]{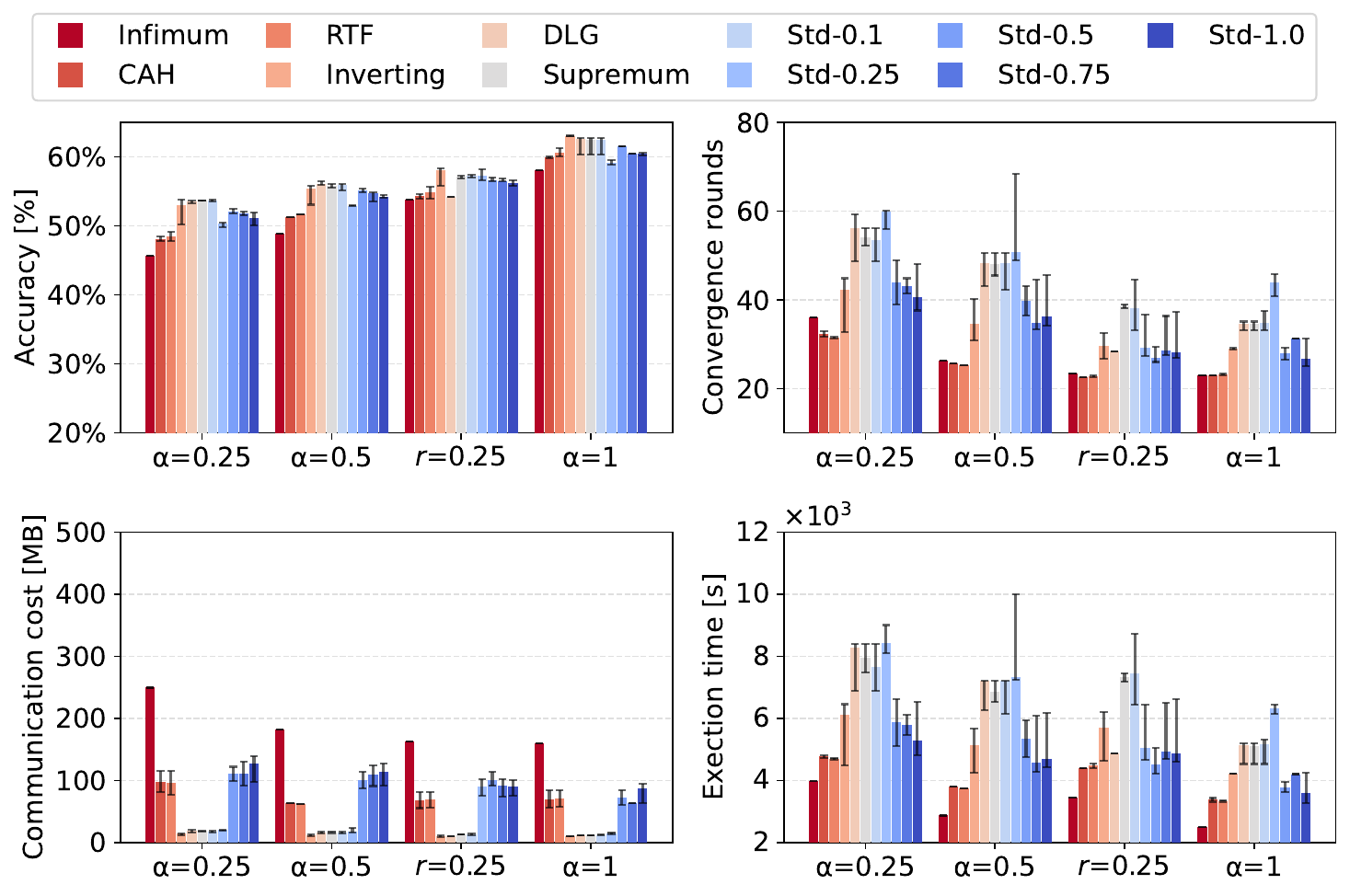}
    \caption{Bar charts of accuracies and system costs for the methods and configurations that meet all of the thresholds $t_{asr} {=} 0.01$, $t_{acc} {=} 0.04$, and $t_{sc} {=} 0.2$.
    For each privacy level and non-IIDness parameter $\alpha$, black error bars indicate the best and worst metrics, while colored bars represent the median values. The  $r {=} 0.25$ corresponds to synthetic data augmentation case.
    }
    \label{fig:RQ4_acc_costs_refined}
\end{figure}

\noindent \textbf{Accuracy and Costs under Privacy Preservation.}
\Cref{fig:RQ4_acc_costs_refined} shows the highest, median, and lowest accuracies and system costs for the configurations with the best trade-offs.
At lower privacy levels (Infimum to RTF), the results resemble \Cref{fig:RQ4_acc_costs}.
In both cases, we observe that communication costs are high, up to 250 MB in total, due to the use of HE, while convergence time and computational cost remain modest (converges at approximately round 35 and 5{,}000 seconds execution time).

From Inverting to Std-0.25, we observe that the accuracy remains high except for Std-0.25, and communication cost remains low.
However, the convergence time and computation cost increase as the privacy level increases.

The main differences emerge at higher privacy levels (Std-0.5 to Std-1.0).
In \Cref{fig:RQ4_acc_costs}, increasing the privacy level leads to lower medium accuracy and higher medium system costs across all $\alpha$ and $r$.
In contrast,  \Cref{fig:RQ4_acc_costs_refined} shows that accuracy remains stable, with minimal degradation, and system costs are better controlled.
For example, at $\alpha {=} 0.5$, Std-1.0 achieves an accuracy of ${\approx}54\%$ versus ${\approx}56\%$ for the best-performing configuration at the same~$\alpha$.

Moreover, the convergence and computational costs at Std-0.5 to Std-1.0 are even lower than those at Std-0.1 to Std-0.25.
Although communication costs still increase at higher privacy levels, they are considerably lower than the worst-case scenario in \Cref{fig:RQ4_acc_costs}, showing that there are methods that achieve better trade-offs at higher privacy levels.
Specifically, at $\alpha {=} 0.5$, the three highest privacy levels (Std-0.5 to Std-1.0) incur around 100 MB of communication cost in \Cref{fig:RQ4_acc_costs_refined}, compared to approximately 400 MB in \Cref{fig:RQ4_acc_costs}.

Overall, strong privacy can be achieved with moderate cost increases and limited accuracy loss.
At intermediate privacy levels, we can achieve high accuracy and low communication cost at the expense of higher convergence time and computational cost.
At the highest levels, accuracy remains high with faster convergence and computational cost but moderately higher communication cost.

\noindent \textbf{Methods for Balancing Accuracy and System Costs.}
We now examine which methods and configurations achieve the best trade-offs between accuracy and system costs.
We derive our findings from \Cref{fig:RQ4_acc_costs_refined}, which shows the results of method configurations that maintain high accuracy without incurring excessive system costs while preserving privacy.
Specifically, we summarize the methods and configurations from \Cref{fig:RQ4_acc_costs_refined} in a table.

\Cref{tab:table_rq4_collapsed} presents the method configurations achieving the best accuracy--cost balance while preserving privacy.
In general, HE-based methods remain essential for RTF, CAH, and Infimum privacy levels.
Specifically, for RTF and CAH, the MP method is particularly effective, achieving  better trade-offs than full encryption ($\eta {=} 1$).
As the noise multipliers increase, DP-based methods become more effective in balancing accuracy and system costs for Inverting to Std-0.25.
At the highest privacy levels (Std-1.0 to Std-0.5), PI provides the best trade-offs while defending against the studied attackers, demonstrating the effectiveness of interleaving DP and HE.
With synthetic augmentation, the selected configurations shift slightly, but the overall trends remain unchanged.

Our privacy budget evaluation in \Cref{sec:noise_mult_to_budget} shows that these privacy levels correspond to privacy budgets~$\varepsilon$ in the range $(10^3, 10^5)$.
Privacy budget substantially larger than $10^5$ (\eg RTF, CAH, and Infimum) provide insufficient protection against the studied attacks using DP alone, while privacy budgets smaller than $10^3$ (\eg Std-0.5 to Std-1.0) significantly degrade learning quality.

\begin{table*}[htbp]
\caption{Methods and their configurations that can balance between accuracy and system costs while preserving privacy (only configurations that meet the thresholds $t_{asr} {=} 0.01$, $t_{acc} {=} 0.04$, and $t_{sc} {=} 0.5$ are included). Methods marked with * are those in which the chosen  configurations  vary by at most two configurations across different privacy levels}
\label{tab:table_rq4_collapsed}
\centering
\begin{tblr}{
  width = \textwidth,
  rowsep = 0.5pt,
  colspec = {X[0.4,l] X[1,l] X[1,l] X[1,l]},
  row{1} = {font=\bfseries},
  hlines, vlines
}
 & {$\alpha = 0.25$} & {$\alpha = 0.5$} & {$\alpha = 1$} \\

{Std-1.0 to \\ Std-0.5} &
{PI\textsuperscript{*}: $\rho \in \{\tfrac{1}{2}, \tfrac{3}{5}, \tfrac{3}{4}\}$} &
{PI\textsuperscript{*}: $\rho \in \{\tfrac{1}{2}, \tfrac{3}{5}, \tfrac{3}{4}\}$} &
{PI\textsuperscript{*}: $\rho \in \{\tfrac{1}{2}, \tfrac{3}{5}, \tfrac{3}{4}\}$} \\

{Std-0.25 \\ Std-0.1} &
{DP (SI/DP, PI, MP)} &
{DP (SI/DP, PI, MP)} &
{SI/DP\textsuperscript{*}: $\rho \in \{0, \tfrac{1}{4}\}$, \\ DP (SI/DP, PI, MP)} \\

Supremum &
{DP (PI, MP)} &
{DP (SI/DP, PI)} &
{SI/DP: $\rho \in \{0, \tfrac{1}{4}\}$,\\ DP (SI/DP, PI, MP)} \\

DLG &
{DP (SI/DP, PI, MP)} &
{DP (SI/DP, PI, MP)} &
{SI/DP: $\rho \in \{0, \tfrac{1}{4}\}$,\\ DP (SI/DP, PI, MP)} \\

Inverting &
{SI/DP: $\rho \in \{0, \tfrac{1}{4}\}$,\\ DP (SI/DP, PI, MP)} &
{SI/DP: $\rho \in \{0, \tfrac{1}{4}\}$,\\ DP (SI/DP, PI, MP)} &
{DP (PI, MP)} \\

RTF/CAH &
{MP: $\eta\in\{0.2,0.4\}$} &
{MP: $\eta\in\{0.2\}$} &
{MP: $\eta\in\{0.2,0.4\}$} \\

Infimum &
{HE (MP)} &
{HE (MP)} &
{HE (MP)} \\
\end{tblr}

\end{table*}

\section{Privacy Preservation vs. System Costs}
\label{sec:privacy_vs_system_costs}

To address RQ3, we analyze attack success rates and system costs under a required accuracy level to identify methods and configurations that achieve the most favorable privacy--cost trade-offs.

\noindent \textbf{Thresholds for Balancing Privacy and System Costs.}
We first consider the method configurations that achieve a specified accuracy level, using the accuracy threshold $t_{acc}$ defined in \Cref{sec:system_costs_vs_learning_quality}.
Specifically, we set $t_{acc} {=} 0.1$ to enable broader exploration.

\begin{figure}[t]
    \centering
    \includegraphics[width=\linewidth]{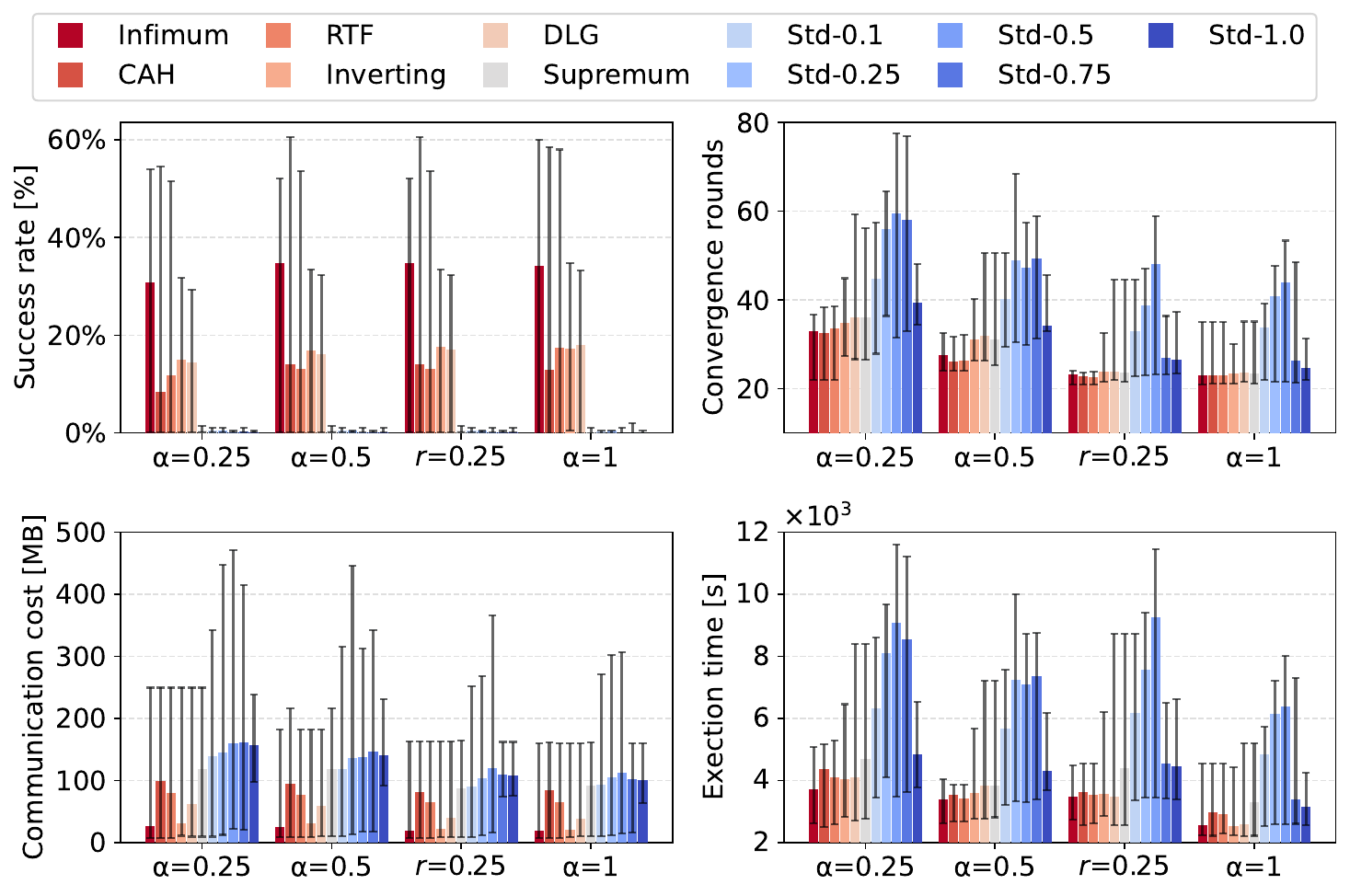}
    \caption{Bar charts of attack success rates and system costs for the methods and configurations that meet the threshold $t_{acc} {=} 0.1$.
    For each privacy level and non-IIDness parameter $\alpha$, black error bars indicate the best and worst metrics, while colored bars represent the median values. The  $r {=} 0.25$ corresponds to synthetic data augmentation case.}
    \label{fig:rq5_succ_costs}
\end{figure}

\Cref{fig:rq5_succ_costs} reports attack success rates and system costs for configurations meeting $t_{acc}$.
As in \Cref{fig:RQ4_acc_costs}, most cases exhibit large error intervals.
Despite this, we observe that the attack success rates decrease significantly for Supremum and higher privacy levels.
Specifically, attack-specific and Infimum privacy levels fail to provide adequate defense, with worst-case attack success rates of 30\%--60\%, whereas the Supremum and standard levels reduce worst-case success rates to 2\%--3\%.
Therefore, only the Supremum or higher privacy levels enable sufficient privacy protection for all methods.

These privacy gains incur higher resource consumption. 
At higher privacy levels (Supremum and higher), we observe increased system costs compared with lower levels (lower than Supremum), except for Std-1.0.
At Std-1.0, the threshold $t_{acc} {=} 0.1$ excludes configurations with both low accuracy and high costs.

To refine our privacy-cost trade-off analysis, we apply $t_{acc}$, $t_{asr}$, and $t_{sc}$ to identify low-cost, privacy-preserving configurations under a required accuracy level.
We use $t_{acc} {=} 0.1$ and $t_{sc} {=} 0.5$, while varying $t_{asr}$ within $[0.001, 0.01]$ in increments of 0.001 to select an appropriate value.
As a result, we use $t_{asr} {=} 0.005$, since higher values do not significantly change the selections.

\begin{figure}[t]
    \centering
    \includegraphics[width=\linewidth]{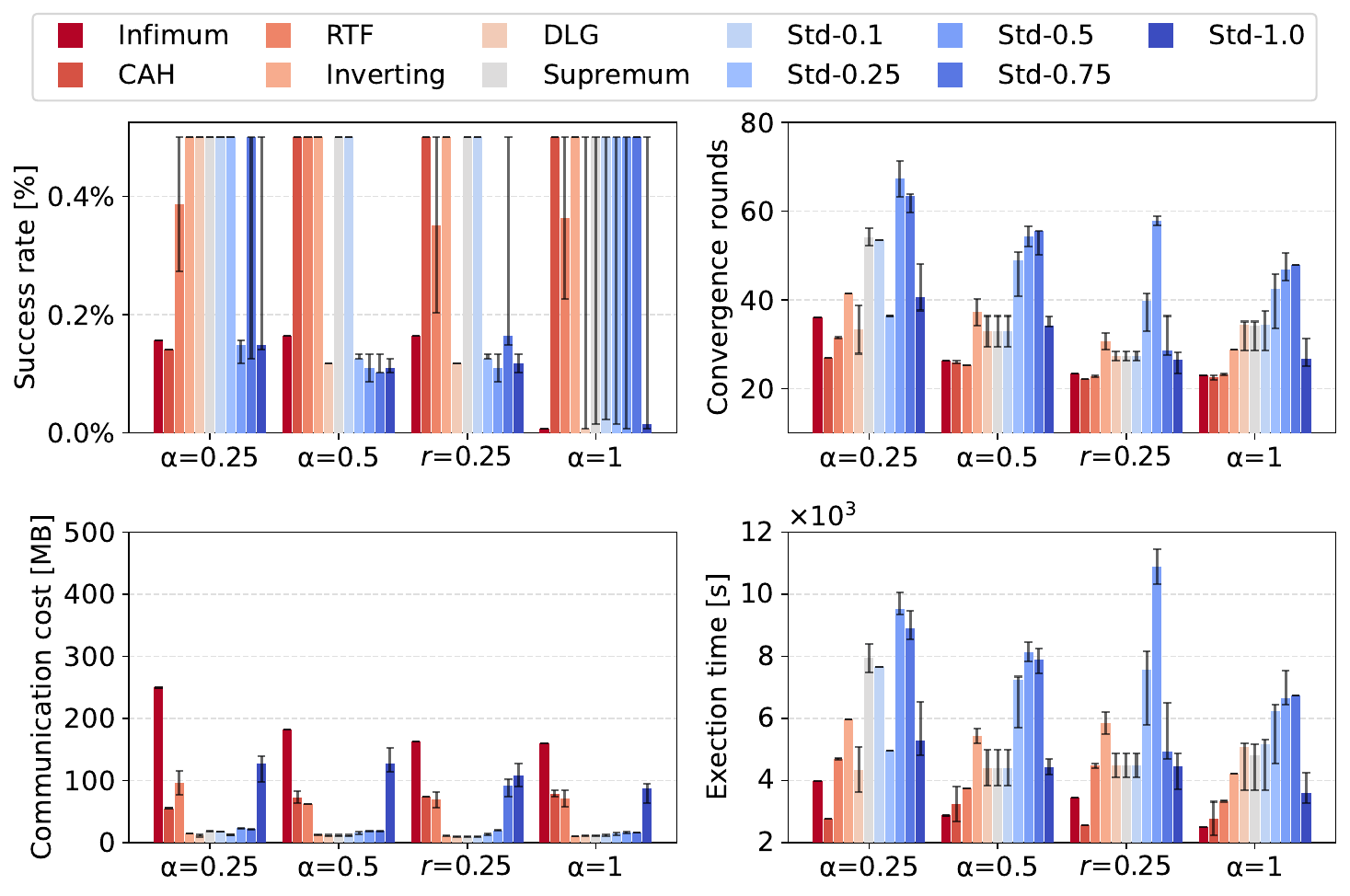}
    \caption{Bar charts of attack success rates and system costs for the methods and configurations that meet the threshold $t_{acc} {=} 0.1$, $t_{asr} {=} 0.005$, and $t_{sc} {=} 0.5 $.
    For each privacy level and non-IIDness parameter $\alpha$, black error bars indicate the best and worst metrics, while colored bars represent the median values. The  $r {=} 0.25$ corresponds to synthetic data augmentation case.}
    \label{fig:rq5_succ_costs_refined}
\end{figure}

\noindent \textbf{Privacy and Costs under Accuracy Relaxation.}
\Cref{fig:rq5_succ_costs_refined} summarizes the success rates and system costs for configurations meeting $t_{acc}{=}0.1$, $t_{asr}{=}0.005$, and $t_{sc}{=}0.5$ across privacy levels and non-IIDness.

From Inverting to Std-0.75, communication costs remain low, indicating that DP-based methods are used in most cases.
Thus, DP alone is sufficient to defend against the studied attacks at these privacy levels.
However, Std-1.0 exhibits a higher communication cost (${\approx}130$MB vs. ${\approx} 20$MB), indicating HE usage and the associated privacy--quality--cost trade-off. 

Overall, lower privacy levels (Infimum, CAH, and RTF) can incur higher communication costs than higher levels due to HE.
For example, at $\alpha {=} 0.5$, Infimum exhibits a communication cost of 180 MB, whereas Std-0.1 to Std-0.75 require at most 20 MB.
Conversely, convergence times and computational costs at lower privacy levels remain lower than those of higher levels (Std-0.1 to Std-0.75), as the DP-only configurations generally converge more slowly and incur higher computational costs.

\begin{table*}[tbp]
\caption{Methods and their configurations that can balance between attack success rates and system costs while maintaining an adequate level of accuracy (only the  configurations meeting the thresholds $t_{acc} = 0.1$, $t_{asr} = 0.005$, and $t_{sc} = 0.5$ are presented). Methods marked with * are those in which the chosen  configurations  vary by at most two configurations across different privacy levels}
\label{tab:table_rq5_collapsed}
\centering
\begin{tblr}{
  width = \textwidth,
  rowsep = 0.5pt,
  colspec = {X[0.4,l] X[1,l] X[1,l] X[1,l]},
  row{1} = {font=\bfseries},
  hlines, vlines
}
 & {$\alpha = 0.25$} & {$\alpha = 0.5$} & {$\alpha = 1$} \\

Std-1.0 &
{PI: $\rho\in\{\tfrac{1}{2}, \tfrac{3}{5}, \frac{3}{4}\}$} &
{PI: $\rho\in\{\tfrac{2}{5}, \tfrac{1}{2}, \frac{3}{5}\}$} &
{PI: $\rho\in\{\tfrac{1}{2}, \tfrac{3}{5}, \frac{3}{4}\}$} \\

{Std-0.75 to Std-0.5} &
{DP (SI/DP, PI, MP)} &
{DP\textsuperscript{*} (SI/DP, PI, MP)} &
{DP\textsuperscript{*} (SI/DP, PI, MP)} \\

Std-0.25 &
{SI/DP: $\rho=\tfrac{1}{4}$} &
{SI/DP: $\rho \in \{0, \tfrac{1}{4}\}$,\\ DP (SI/DP, PI)} &
{SI/DP: $\rho=\tfrac{1}{4}$,\\ DP (SI/DP, PI, MP)} \\

Std-0.1 &
{DP (PI)} &
{SI/DP: $\rho\in\{\tfrac{1}{4}, \tfrac{2}{5}\}$} &
{SI/DP: $\rho\in\{\tfrac{1}{4}, \tfrac{2}{5}\}$,\\ DP (SI/DP, PI, MP)} \\

Supremum &
{DP (PI, MP)} &
{SI/DP: $\rho\in\{\tfrac{1}{4}, \tfrac{2}{5}\}$} &
{SI/DP: $\rho\in\{\tfrac{1}{4}, \tfrac{2}{5}\}$,\\ DP (SI/DP, PI)} \\

DLG &
{SI/DP: $\rho\in\{\tfrac{1}{4}, \tfrac{2}{5}\}$} &
{SI/DP: $\rho\in\{\tfrac{1}{4}, \tfrac{2}{5}\}$} &
{SI/DP: $\rho\in\{\tfrac{1}{4}, \tfrac{2}{5}\}$,\\ DP (SI/DP, PI, MP)} \\

Inverting &
{DP (MP)} &
{DP (PI, MP)} &
{DP (MP)} \\

{RTF} &
{MP: $\eta\in\{0.2, 0.4\}$} &
{MP: $\eta=0.2$} &
{MP: $\eta\in\{0.2, 0.4\}$} \\

{CAH} &
{SI/HE: $\rho=\tfrac{3}{4}$} &
{SI/HE: $\rho=\tfrac{3}{5}$,\\ MP: $\eta=0.2$} &
{SI/HE: $\rho=\tfrac{3}{5}$,\\ MP: $\eta=0.4$} \\

Infimum &
{HE (MP)} &
{HE (MP)} &
{HE (MP)} \\
\end{tblr}
\end{table*}

\noindent \textbf{Methods for Balancing Privacy and System Costs.}
We now examine which methods and configurations provide the most effective trade-offs between privacy and system costs under a required accuracy level. 
Following our approach in \Cref{sec:system_costs_vs_learning_quality}, we identify the configurations with the best privacy--cost balance in \Cref{tab:table_rq5_collapsed}.

At Std-1.0, PI provides the best privacy--cost across all $\alpha$ values while maintaining acceptable accuracy.
At intermediate levels (Inverting to Std-0.75), DP-based methods dominate the selected configurations, as these methods achieve significantly lower communication costs while still providing adequate protection.
Compared to \Cref{sec:system_costs_vs_learning_quality}, DP-based configuration selections extend to higher levels (Std-0.5 and higher) because the relaxed accuracy requirement enables further cost reductions.
At weaker privacy levels (Infimum, RTF, and CAH), HE becomes necessary again, since the noise multipliers at these levels are too weak for DP alone to provide sufficient defense, similar to the results in \Cref{tab:table_rq4_collapsed}.

Results with synthetic augmentation largely mirror those in the non-augmented case.
For example, at Std-1.0, PI is selected with $\rho{\in}\{\tfrac{1}{4}, \tfrac{2}{5}, \tfrac{1}{2}, \tfrac{3}{5}\}$.
The main difference in the selected methods occurs at Std-0.75, where PI is selected instead of DP-only because DP-only significantly degrades the accuracy when augmentation is used, making PI the most suitable option.

\section{Evaluation on the Fashion-MNIST Dataset} \label{sec:fashionmnist}

\begin{table}[tbp]
\caption{DP noise multipliers $\sigma$ selected for each encryption ratio $\eta$ for all privacy levels (PL) using Fashion-MNIST} 
\label{tab:privacy_level_params_fmnist}
\centering
\begin{tblr}{
  width=\columnwidth,
  colspec = {l|llllllll},
  colsep = 1.8pt,
  rowsep = 0.5pt,
  row{1} = {font=\bfseries},
  column{1} = {font=\bfseries},
}
\toprule
\diagbox{PL}{$\eta$} & 0 & 0.05 & 0.2 &  0.4 & 0.6 & 0.7 & 0.8 & 1 \\
\midrule

Std-1.0 & \SetCell[c=7]{c} 1.0 &  &  &   &  &  &  &  \\
Std-0.75 & \SetCell[c=7]{c} 0.75 &  &  &  &  &  &  &  \\
Std-0.5 & \SetCell[c=7]{c} 0.5 &  &  &  &  &  &  &  \\
Std-0.25 & \SetCell[c=7]{c} 0.25 &  &  &  &  &  &  &  \\
Std-0.1 & \SetCell[c=7]{c} 0.1 &  &  &  &  &  &  &  \\
\midrule
Supremum & 0.05 & 0.001 & 0.001 & 0.001 & 0.005 & 0.001 & 0.0005 & \SetCell{cmd=\fbox} 0 \\
DLG & 0.05 & \SetCell{cmd=\fbox} 0 &  0 & 0 & 0 & 0 & 0 & 0 \\
Inverting & 0.05 & \SetCell{cmd=\fbox} 0 & 0 & 0 & 0 & 0 & 0 & 0 \\
RTF & 0.00005 & 0.00005 & 0.00005  & 0.001 & 0.005 & \SetCell{cmd=\fbox} 0 & 0 & 0 \\
CAH & 0.001 & 0.001 & 0.001 & 0.001  & 0.001 & 0.001 & 0.0005  & \SetCell{cmd=\fbox} 0 \\
Infimum &  0.00005 & \SetCell{cmd=\fbox} 0 & 0 & 0 & 0 & 0 & 0 & 0 \\
\bottomrule

\end{tblr}
\end{table}

\begin{figure}[tbp]
    \centering
    \includegraphics[width=\linewidth]{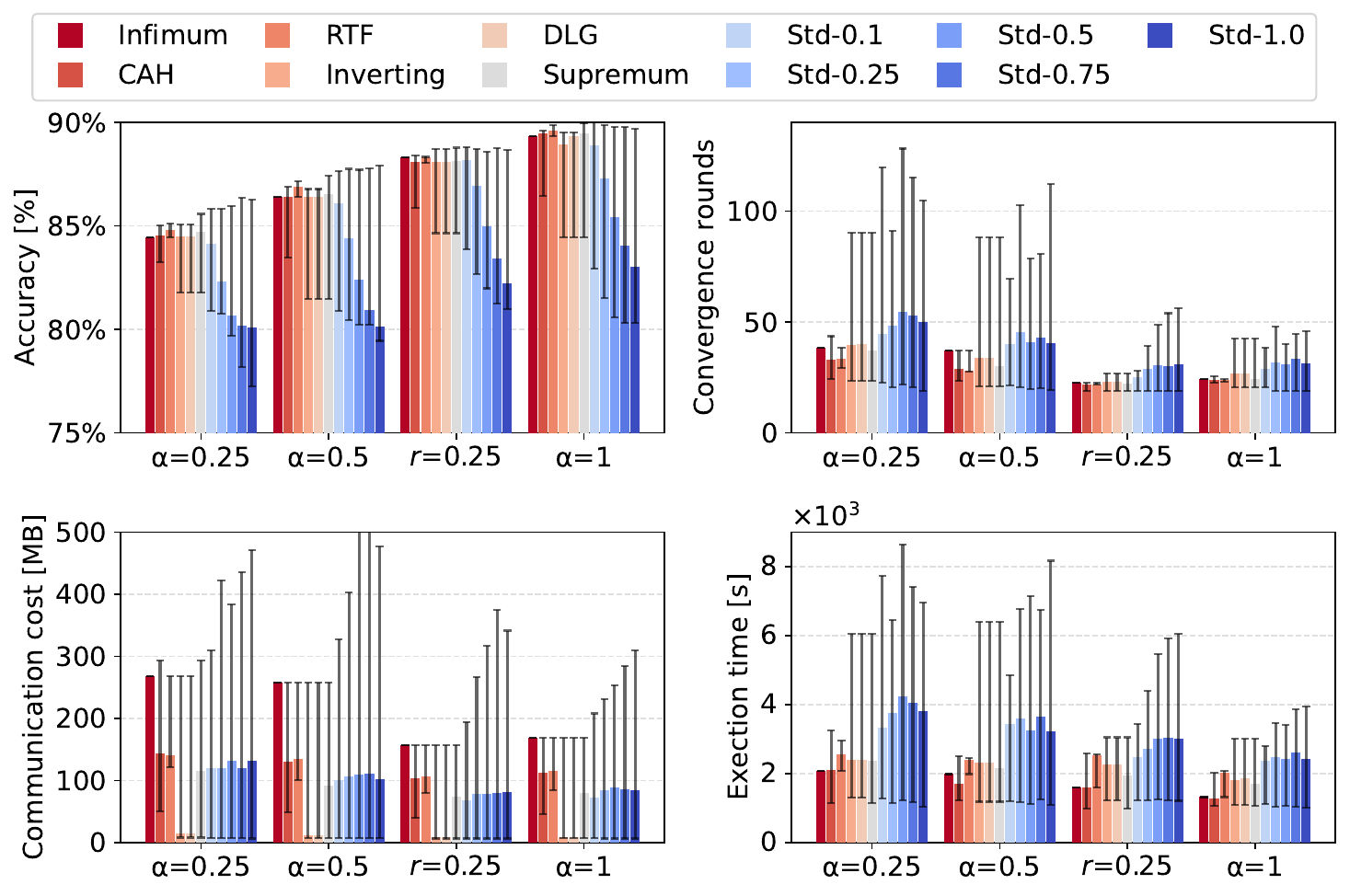}
    \caption{Bar charts of accuracies and system costs for the methods and configurations that meet the threshold $t_{asr} {=} 0.01$ when trained on Fashion-MNIST.
    For each privacy level and non-IIDness parameter $\alpha$, black error bars indicate the best and worst metrics, while colored bars represent the median values. The $r {=} 0.25$ corresponds to the synthetic data augmentation case.
    }
    \label{fig:RQ4_acc_costs_fmnist}
\end{figure}

\begin{table*}[tbp]
\caption{Methods and their configurations that provide the best defenses against the privacy attacks under varying data distribution, with $t_{asr}=0.5\%$ using Fashion-MNIST. Note that the case with synthetic augmentation is identical to $\alpha {=} 0.5$. Methods marked with $*$  are  those in which the chosen configurations vary by at most two configurations across different privacy levels}
\label{tab:best_privacy_collapsed_fmnist}
\centering
\begin{tblr}{
  width = \textwidth,
  rowsep = 0.5pt,
  colspec = {X[0.4,l] X[1,l] X[1,l] X[1,l]},
  row{1} = {font=\bfseries},
  hlines, vlines
}
 & {$\alpha = 0.25$} & {$\alpha = 0.5$} & {$\alpha = 1$} \\

Std-1.0 to Std-0.1 &
{SI/DP, PI\textsuperscript{*}, MP\textsuperscript{*}} &
{SI/DP, PI\textsuperscript{*}, MP\textsuperscript{*}} &
{SI/DP, PI\textsuperscript{*}, MP\textsuperscript{*}} \\

Supremum &
{SI/HE, PI, MP (excl. $\eta=1$)} &
{SI/DP, SI/HE,\\ PI (excl. $\rho\in\{\tfrac{1}{2}, \tfrac{3}{5}\}$), \\ MP} & 
{SI/HE, \\ PI (excl. $\rho\in\{\tfrac{1}{2}, \tfrac{3}{5}\}$),\\ MP} \\

{DLG/Inverting} &
{SI/DP,\\ DP\textsuperscript{*} (SI/DP, PI, MP), HE (MP)} &
{SI/DP,\\ DP\textsuperscript{*} (SI/DP, PI, MP), HE (MP)} &
{SI/DP,\\ DP\textsuperscript{*} (SI/DP, PI, MP), HE (MP)} \\

RTF &
{MP: $\eta\in\{0.4, 0.6, 1\}$ \\ HE (MP)} &
{MP: $\eta\in\{0.4, 0.6, 1\}$ \\ HE (MP)} &
{MP: $\eta\in\{0.4, 0.6, 1\}$ \\ HE (MP)} \\

CAH &
{SI/HE, \\ MP (excl. $\eta=0$), \\ HE (MP, PI, SI/HE) } &
{SI/HE, \\ MP (excl. $\eta\in\{0, 0.4\}$), \\ HE (MP, PI, SI/HE) } &
{SI/HE, \\ MP (excl. $\eta=0$), \\ HE (MP, PI, SI/HE) } \\

Infimum &
{HE (MP)} &
{HE (MP)} &
{HE (MP)} \\
\end{tblr}

\end{table*}

\begin{table*}[tbp]
\caption{Methods and their configurations that can balance between accuracy and system costs while preserving privacy using Fashion-MNIST (only configurations that meet the thresholds $t_{asr} {=} 0.01$, $t_{acc} {=} 0.04$, and $t_{sc} {=} 0.5$ are included). Methods marked with * are those in which the chosen  configurations  vary by at most two configurations across different privacy levels}
\label{tab:table_rq4_collapsed_fmnist}
\centering
\begin{tblr}{
  width = \textwidth,
  rowsep = 0.5pt,
  colspec = {X[0.4,l] X[1,l] X[1,l] X[1,l]},
  row{1} = {font=\bfseries},
  hlines, vlines
}
 & {$\alpha = 0.25$} & {$\alpha = 0.5$} & {$\alpha = 1$} \\

{Std-1.0 to \\ Std-0.5} &
{PI\textsuperscript{*}: $\rho \in \{\tfrac{1}{2}, \tfrac{3}{5}, \tfrac{3}{4}\}$} &
{PI\textsuperscript{*}: $\rho \in \{\tfrac{1}{2}, \tfrac{3}{5}, \tfrac{3}{4}\}$} &
{PI\textsuperscript{*}: $\rho \in \{\tfrac{1}{2}, \tfrac{3}{5}, \tfrac{3}{4}\}$} \\

{Std-0.25} &
{DP (SI/DP, PI, MP)} &
{DP (SI/DP, PI, MP)} &
{SI/DP: $\rho \in \{0, \tfrac{1}{4}\}$, \\ DP (SI/DP, PI, MP)} \\

{Std-0.1} &
{SI/DP: $\rho\in\{\tfrac{2}{5}, \tfrac{1}{2}\}$} &
{SI/DP: $\rho = \tfrac{2}{5}$,\\ DP (PI)} &
{SI/DP: $\rho \in \{0, \tfrac{2}{5}, \tfrac{1}{2}\}$, \\ DP (SI/DP, PI, MP)} \\

{Supremum} &
{SI/DP: $\rho\in\{\tfrac{3}{5}, \tfrac{3}{4}\}$} &
{SI/DP: $\rho \in \{0, \tfrac{2}{5}, \tfrac{1}{2}\}$, \\ DP (SI/DP, MP)} &
{SI/DP: $\rho \in \{0, \tfrac{2}{5}, \tfrac{1}{2}\}$} \\

{DLG/Inverting} &
{SI/DP: $\rho \in \{\tfrac{3}{5}, \tfrac{3}{4}\}$} &
{SI/DP: $\rho \in \{\tfrac{2}{5}, \tfrac{1}{2}, \tfrac{3}{5}\}$} &
{SI/DP: $\rho \in \{\tfrac{2}{5}, \tfrac{1}{2}, \tfrac{3}{5}\}$} \\

RTF &
{MP: $\eta\in\{0.4,0.6\}$} &
{MP: $\eta\in\{0.4,0.6\}$} &
{MP: $\eta\in\{0.4,0.6\}$} \\

CAH &
{SI/HE: $\rho=\tfrac{3}{4}$} &
{SI/HE: $\rho=\tfrac{3}{4}$} &
{SI/HE: $\rho=\tfrac{3}{4}$} \\

Infimum &
{HE (MP)} &
{HE (MP)} &
{HE (MP)} \\
\end{tblr}

\end{table*}

\begin{table*}[tbp]
\caption{Methods and their configurations that can balance between attack success rates and system costs while maintaining an adequate level of accuracy using Fashion-MNIST (only the configurations meeting the thresholds $t_{acc} = 0.1$, $t_{asr} = 0.005$, and $t_{sc} = 0.5$ are presented). Methods marked with * are those in which the chosen  configurations  vary by at most two configurations across different privacy levels}
\label{tab:table_rq5_collapsed_fmnist}
\centering
\begin{tblr}{
  width = \textwidth,
  rowsep = 0.5pt,
  colspec = {X[0.4,l] X[1,l] X[1,l] X[1,l]},
  row{1} = {font=\bfseries},
  hlines, vlines
}
 & {$\alpha = 0.25$} & {$\alpha = 0.5$} & {$\alpha = 1$} \\

{Std-1.0 to Std-0.1} &
{SI/DP\textsuperscript{*}: $\rho\in\{\tfrac{1}{2}, \tfrac{3}{5}, \frac{3}{4}\}$} &
{SI/DP\textsuperscript{*}: $\rho\in\{\tfrac{1}{2}, \tfrac{3}{5}, \frac{3}{4}\}$} &
{SI/DP\textsuperscript{*}:
: $\rho\in\{\tfrac{1}{2}, \tfrac{3}{5}, \frac{3}{4}\}$} \\

Supremum &
{DP (PI)} &
{SI/DP: $\rho\in\{\tfrac{3}{5}, \tfrac{3}{4}\}$} &
{DP (MP)} \\

DLG/Inverting &
{SI/DP: $\rho\in\{\tfrac{3}{5}, \tfrac{3}{4}\}$} &
{SI/DP: $\rho\in\{\tfrac{3}{5}, \tfrac{3}{4}\}$} &
{SI/DP: $\rho\in\{\tfrac{3}{5}, \tfrac{3}{4}\}$} \\

{RTF} &
{MP: $\eta\in\{0.4, 0.6\}$} &
{MP: $\eta\in\{0.4, 0.6\}$} &
{MP: $\eta\in\{0.4, 0.6\}$} \\

{CAH} &
{SI/HE: $\rho=\tfrac{3}{4}$} &
{SI/HE: $\rho=\tfrac{3}{4}$} &
{SI/HE: $\rho=\tfrac{3}{4}$} \\

Infimum &
{HE (MP)} &
{HE (MP)} &
{HE (MP)} \\
\end{tblr}
\end{table*}

To strengthen our empirical analysis, we also evaluate the Fashion-MNIST dataset~\cite{DBLP:journals/corr/abs-1708-07747}, presenting the results in the summary tables (\Cref{tab:best_privacy_collapsed_fmnist,tab:table_rq4_collapsed_fmnist,tab:table_rq5_collapsed_fmnist}), as the overall conclusions remain largely consistent.
Fashion-MNIST consists of 60,000 grayscale training images spanning 10 categories of fashion products. Compared to CIFAR-10~\cite{krizhevsky2009learning}, it presents a slightly larger dataset and represents a somewhat easier classification task due to the simpler visual structure of the images.

\Cref{tab:best_privacy_collapsed_fmnist} reports the resulting noise multiplier $\sigma$ for each encryption ratio $\eta$, obtained by repeating the privacy parameter evaluation procedure described in \Cref{sec:privacy_levels} on Fashion-MNIST.
Compared to \Cref{tab:best_privacy_collapsed}, which uses CIFAR-10, the privacy levels for Fashion-MNIST generally require lower privacy parameters, indicating that Fashion-MNIST is easier to protect.
Additionally, the DLG and Inverting privacy levels yield identical privacy parameters.
Consequently, their results are expected to be similar.
Using the derived privacy levels, we repeat the full experimental evaluation.

The Fashion-MNIST results for each research question are presented in \Cref{tab:best_privacy_collapsed_fmnist,tab:table_rq4_collapsed_fmnist,tab:table_rq5_collapsed_fmnist}.
Compared to CIFAR-10, the differences are minor but worth noting.
First, for Fashion-MNIST, the results for DLG and Inverting privacy levels are consistently merged, as both select similar methods and configurations.
This does not alter the overall trend: for both datasets, DLG and Inverting levels favor DP-based approaches (SI/DP and DP-only).
Second, SI/DP is more frequently selected than DP-only for Fashion-MNIST, whereas the opposite holds for CIFAR-10 (see \Cref{tab:table_rq4_collapsed_fmnist,tab:table_rq5_collapsed_fmnist}).
Additionally, in \Cref{tab:table_rq5_collapsed_fmnist}, SI/DP is selected instead of PI at Std-1.0, unlike in \Cref{tab:table_rq5_collapsed} when CIFAR-10 is used. 
This difference arises because DP and synthetic data introduce a smaller degradation in learning quality for Fashion-MNIST.

This effect is evident by inspecting the accuracy distributions.
For CIFAR-10 (\Cref{fig:RQ4_acc_costs}), accuracy ranges approximately from  $25\%$ to $63\%$ across configurations.
In contrast, for Fashion-MNIST (\Cref{fig:RQ4_acc_costs_fmnist}), accuracy ranges approximately from $77\%$ to $90\%$, representing a substantially narrower spread.
This smaller variance indicates that privacy protection mechanisms and synthetic data have a reduced impact on learning quality for Fashion-MNIST.

As a result, SI/DP emerges as a better alternative to DP-only and, under relaxed accuracy requirements, to PI, providing acceptable accuracy at lower system costs.
However, under stricter accuracy constraints, such as $t_{acc}{=}0.004$ in \Cref{tab:table_rq4_collapsed_fmnist}, PI remains more suitable at higher privacy levels.
Overall, the trends are consistent with those observed for CIFAR-10, and the method selection process depicted in \Cref{fig:methods_high_level_objective} therefore applies to Fashion-MNIST as well.

\section{Summary of Results and Discussion}
\label{sec:discussion}

\begin{figure}[tbp]
    \centering
    \includegraphics[width=\linewidth]{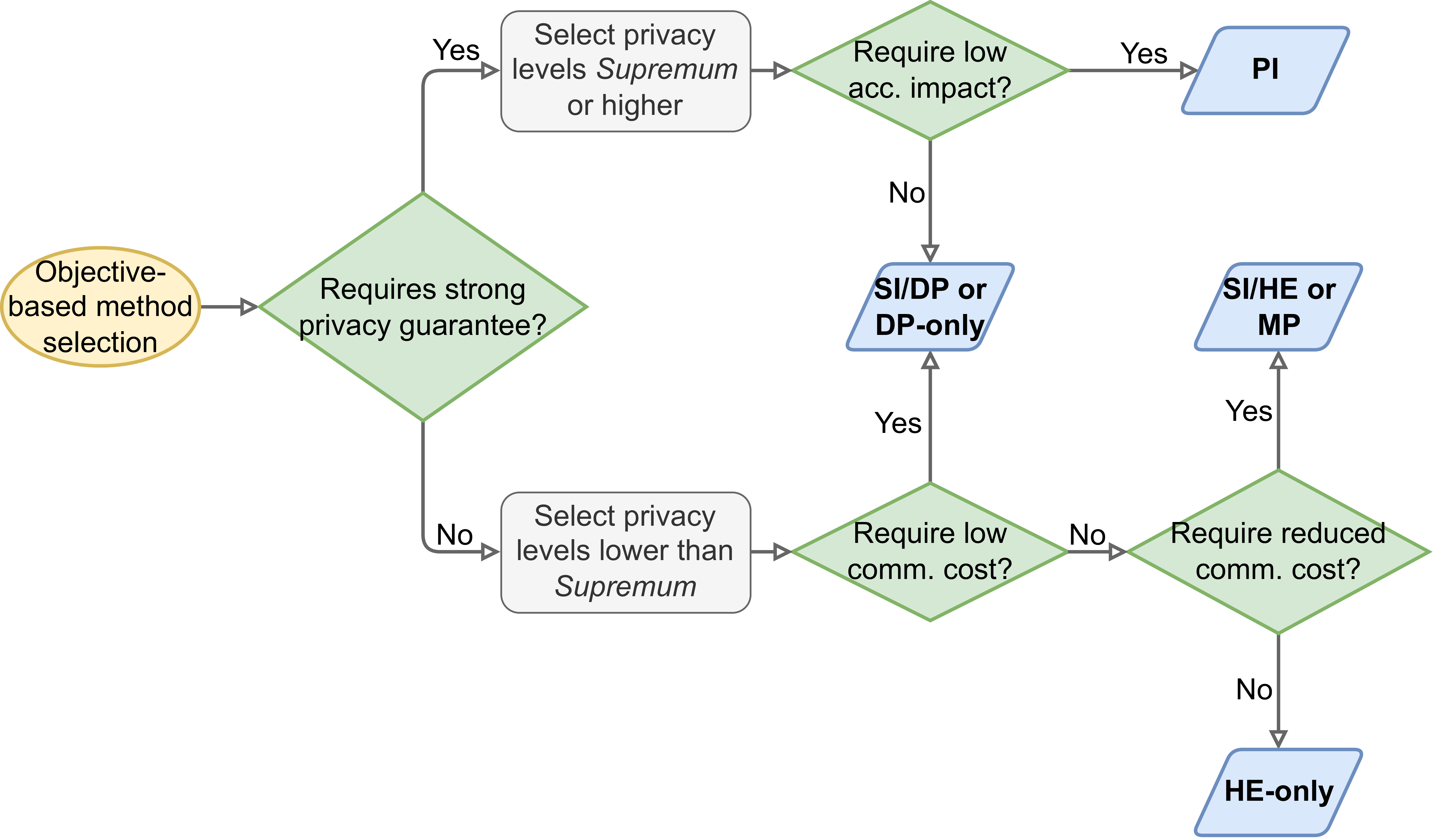}
    \caption{Objective-oriented selection process of the studied methods (represented as blue parallelograms) based on the recommendations derived from our results. The green rhombuses show the prioritized criteria.}
    \label{fig:methods_high_level_objective}
\end{figure}

Based on our empirical findings for the studied datasets and model, this section discusses  
an objective-oriented method-selection process  (depicted in \Cref{fig:methods_high_level_objective})  that considers privacy, accuracy, and communication costs.
The process begins by selecting the desired privacy guarantee.
If a strong privacy guarantee is required (Supremum or higher privacy levels), most methods can defend against the studied attacks, as shown in \Cref{tab:best_privacy_collapsed}.

Under a strong privacy guarantee, we next consider the accuracy requirement.
If high accuracy is required, we select PI because it is the only method that mitigates accuracy degradation caused by strong DP noise, as demonstrated in \Cref{tab:table_rq4_collapsed,tab:table_rq5_collapsed}. 
Otherwise, SI/DP or DP-only can be used because they incur the lowest communication costs.

If a strong privacy guarantee is not required (but some privacy is still desired), we consider the remaining privacy levels, \ie, those lower than Supremum.
To achieve low communication cost, the options are SI/DP and DP-only, as these methods do not employ S-HE.
However, careful selection of the DP noise multiplier is necessary, since only the DLG privacy level can provide sufficient protection using SI/DP or DP-only, as shown in \Cref{tab:table_rq4_collapsed,tab:table_rq5_collapsed}.
The remaining methods employ S-HE for privacy protection.
Hence, the next decision is whether MP or SI/HE should be used to reduce HE-induced communication costs.

DP is a widely adopted technique for protecting FL systems.
As shown in \Cref{fig:methods_high_level_objective}, DP (whether used as an SI/DP or DP-only method) provides an attractive solution in scenarios where low communication cost is the main requirement.
However, DP alone does not solve all the challenges of providing privacy in FL.
In cases where a strong privacy guarantee is required, our results demonstrate that DP can significantly degrade model accuracy.
Consequently, when accuracy is a priority under the strong privacy guarantee requirement, DP becomes less suitable than the proposed PI solution.

\section{Conclusions}

We enhanced private FL by introducing and comparing several methods that integrate FedAvg with DP and S-HE to address key challenges in FL, namely, safeguarding privacy while mitigating the accuracy loss  associated with DP and overhead associated with S-HE.
These enhancements leverage an interleaving strategy that alternates between  training rounds that use DP and S-HE, while reducing the costs of privacy protection via synthetic data.
When synthetic data is used, the models are transmitted in plaintext.

As future work, we encourage the community to build upon our proposed interleaving methods, empirical evaluation framework, and selection procedure.
By leveraging the implementation, which we pledge to release upon the acceptance of this paper, researchers can extend our study to additional datasets beyond CIFAR-10 and Fashion-MNIST, additional model architectures, and diverse deployment scenarios, further generalizing and stress-testing the proposed privacy–quality–efficiency tradeoff analysis.

\bibliographystyle{IEEEtran}
\bibliography{ref}

\cleardoublepage
\begin{appendices}

\section{Glossary and Notation}
\label{app:glossary}

\subsection{Abbreviations}
\label{app:abbr}

\begin{description}[font=\bfseries, labelwidth=1.5cm, leftmargin=2.0cm, align=left]
    \item[Alt-FL] Alternating Federated Learning
    \item[CAH] When the Curious Abandon Honesty
    \item[DLG] Deep Leakage from Gradients
    \item[DP] Differential Privacy
    \item[DP-SGD] Differentially Private Stochastic Gradient Descent
    \item[FL] Federated Learning
    \item[HE] Homomorphic Encryption
    \item[IIP] Image Identifiability Precision
    \item[Inverting] Inverting Gradients
    \item[MP] Mixed Protections
    \item[PI] Privacy Interleaving
    \item[RDP] R\'enyi Differential Privacy
    \item[RTF] Robbing the Fed
    \item[S-HE] Selective Homomorphic Encryption
    \item[SI/DP] Synthetic Interleaving with Differential Privacy
    \item[SI/HE] Synthetic Interleaving with Homomorphic Encryption
\end{description}

\subsection{Mathematical Notation}
\label{app:notation}

\begin{table}[htbp]
\centering
\begin{tblr}{
  colspec = {l X},
  rowsep = 0.5pt,
  column{1} = {halign=c},
  row{1} = {font=\bfseries, halign=c},
}
\toprule
Symbol & Meaning \\
\midrule
$\rho$ & Interleaving ratio. \\
$\eta$ & Encryption ratio for S-HE.\\
$\sigma$ & Noise multiplier parameter in DP mechanisms. \\
$C$ & Maximum gradient clipping norm. \\
$\alpha$ & Non-IIDness parameter for Dirichlet distribution. \\
$r$ & Synthetic data augmentation ratio. \\
$w_i^t$ & Model parameters; subscript $t$ denotes round and superscript $i$ denotes client. \\
$d_{i,a}$ & Authentic data for client $i$. \\
$d_{i,s}$ & Synthetic data for client $i$. \\
$M$ & Encryption mask. \\
$\varepsilon$ & Privacy budget parameter in DP controlling privacy loss. \\
$\delta$ & Probability parameter in $(\varepsilon, \delta)$-DP. \\
\bottomrule
\end{tblr}
\end{table}

\end{appendices}
\end{document}